\documentclass[runningheads]{llncs}

\usepackage{eccv}

\usepackage{eccvabbrv}

\usepackage{wrapfig}
\usepackage{graphicx}
\usepackage{subcaption} % 用于子图排版
\usepackage{xcolor}
\usepackage[table]{xcolor}
\usepackage{multirow}
\usepackage{booktabs}
\usepackage{amsmath}
\usepackage{amssymb}
\usepackage{subfiles}
\usepackage[accsupp]{axessibility}  % Improves PDF readability for those with disabilities.
\usepackage{hyperref}

\usepackage{orcidlink}
\definecolor{mplblue}{RGB}{31, 119, 180}  % Matplotlib 默认蓝色
\definecolor{mplorange}{RGB}{255, 127, 14} % Matplotlib 默认橙色
\definecolor{mplgreen}{RGB}{44, 160, 44}   % Matplotlib 默认绿色
\definecolor{mplred}{RGB}{214, 39, 40}     % Matplotlib 默认红色

\begin{document}
\title{GeoFlow: Efficient Driving Video Generation via Geometry-Aligned Priors} 
% \title{Supplementary Material for GeoFlow} 

\titlerunning{GeoFlow}

% Jiazheng Liu, Hang Li, Jiawei Zhang, Jiahe Li, Xiaohan Yu, Shengyin Fan, Jin Zheng, Xiao Bai
\author{Jiazheng Liu\inst{1} \and
Hang Li\inst{1} \and
Jiawei Zhang\inst{1} \and
Jiahe Li\inst{1} \and
Xiaohan Yu\inst{5} \and
Shengyin Fan\inst{6} \and
Jin Zheng\inst{1,2,4}\textsuperscript{*} \and
Xiao Bai\inst{1,3,4}\textsuperscript{*}
}
% \author{Jiazheng Liu\inst{1}\orcidlink{0009-0008-0022-799X} \and
% Hang Li\inst{1}\orcidlink{1111-2222-3333-4444} \and
% Jiawei Zhang\inst{1}\orcidlink{0009-0003-6321-1593} \and
% Jiahe Li\inst{1}\orcidlink{0009-0007-8669-921X} \and
% Xiaohan Yu\inst{3}\orcidlink{0000-0001-6186-0520} \and
% Shengyin Fan\inst{4}\orcidlink{0009-0000-8611-0339} \and
% Jin Zheng\inst{1,2}\thanks{Corresponding author: Jin Zheng (jinzheng@buaa.edu.cn).} \and
% Xiao Bai\inst{1} 
% }

\authorrunning{J. Liu et al.}

\institute{School of Computer Science and Engineering, Beihang University \and
State Key Laboratory of Virtual Reality Technology and Systems, Beijing \and
State Key Laboratory of Complex \&  Critical Software Environment, Beijing \and
Jiangxi Research Institute, Beihang University \and
Macquarie University \and
Tianyi Transportation Technology Co., Ltd, Suzhou 215133, China
% \email{\{liujiazheng, jinzheng\}@buaa.edu.cn}
}

\maketitle
\renewcommand{\thefootnote}{\fnsymbol{footnote}}
\footnotetext[1]{Corresponding authors: Jin Zheng (jinzheng@buaa.edu.cn) and Xiao Bai (baixiao@buaa.edu.cn).}

\begin{abstract}
% While recent generative models have demonstrated remarkable capability in synthesizing high-fidelity driving videos, they are severely constrained by slow inference speeds due to the requirement of extensive sampling steps.
% In particular, state-of-the-art approaches typically rely on a standard Gaussian source distribution.
% This "starting from scratch" paradigm ignores the rich spatiotemporal correlations inherent in driving scenarios, thereby not only incurring computational redundancy but also struggling to maintain geometric consistency, especially in few-step generation.
% In this paper, we propose a novel framework that accelerates driving video generation by harnessing the inherent structural regularity of driving scenarios to construct Structure-Aware Initial Priors.
% Specifically, we leverage multi-view geometry to build an informative source distribution that aligns closely with the target data manifold.
% This formulation avoids sampling from pure noise, thus bridging the distributional gap between the source and the target, yielding a significantly straighter and shorter sampling trajectory without introducing any extra learnable parameters.
% Extensive experiments demonstrate that our method significantly mitigates temporal drift in few-step generation, achieving state-of-the-art visual quality in just 8 inference steps, matching the performance of baselines at 40 steps.

Generative models like Diffusion Models and Flow Matching have demonstrated remarkable capabilities in synthesizing high-fidelity driving videos, but are severely constrained by high inference latency due to the requirement of extensive sampling steps.
We argue that this inefficiency stems from the prevailing reliance on a standard Gaussian source distribution, where consecutive frames are initialized as independent Gaussian noise.
This paradigm disregards the rich spatiotemporal correlations inherent in driving videos, compelling the model to regenerate deterministic scene structures existing in previous frames from noise, which is both computationally redundant and prone to geometric inconsistency.
To address this problem, we propose GeoFlow, a novel framework designed to achieve efficient driving video generation by harnessing explicit geometric priors.
Instead of sampling from standard Gaussian noise, we leverage multi-view geometry and spatially-adaptive noise injection to construct a Geometry-Aligned Prior (GAP) distribution as starting point. 
This initialization bridges the gap between source distribution and data distribution, yielding a significantly straighter and shorter sampling trajectory.
Extensive experiments demonstrate that GeoFlow can achieve remarkable efficiency of both training and inference: 
merely several hours of fine-tuning on baseline models can significantly boost few-step generation quality, while fully converged training 
drastically reduces number of inference steps required for state-of-the-art video generation.
% Specifically, we leverage multi-view geometry and design an adaptive nosie injection strategy to construct a Geometry-Aligned Prior (GAP) that aligns closely with the target data manifold.
% unlocks a superior visual performance ceiling to the original baselines.
% By fine-tuning existing driving video models, GeoFlow demonstrates remarkable efficiency: merely hundreds of update steps are sufficient to significantly boost few-step generation quality, while full optimization unlocks a visual performance ceiling superior to the original baselines.

\end{abstract}

\begin{figure*}[t]
    \centering
    % ==========================================
    % 左侧列 (Left Column): 占 40% 宽度
    % 包含 (a) 原理图 和 (b) 效率曲线
    % ==========================================
    \begin{minipage}[b]{0.32\linewidth}
        % --- Subfigure (a): Top Left ---
        \begin{subfigure}[b]{\linewidth}
            \centering
            % [TODO] 替换为原理图 (Schematic)
            % 建议比例: 4:3
            \includegraphics[width=\linewidth, height=3.4cm]{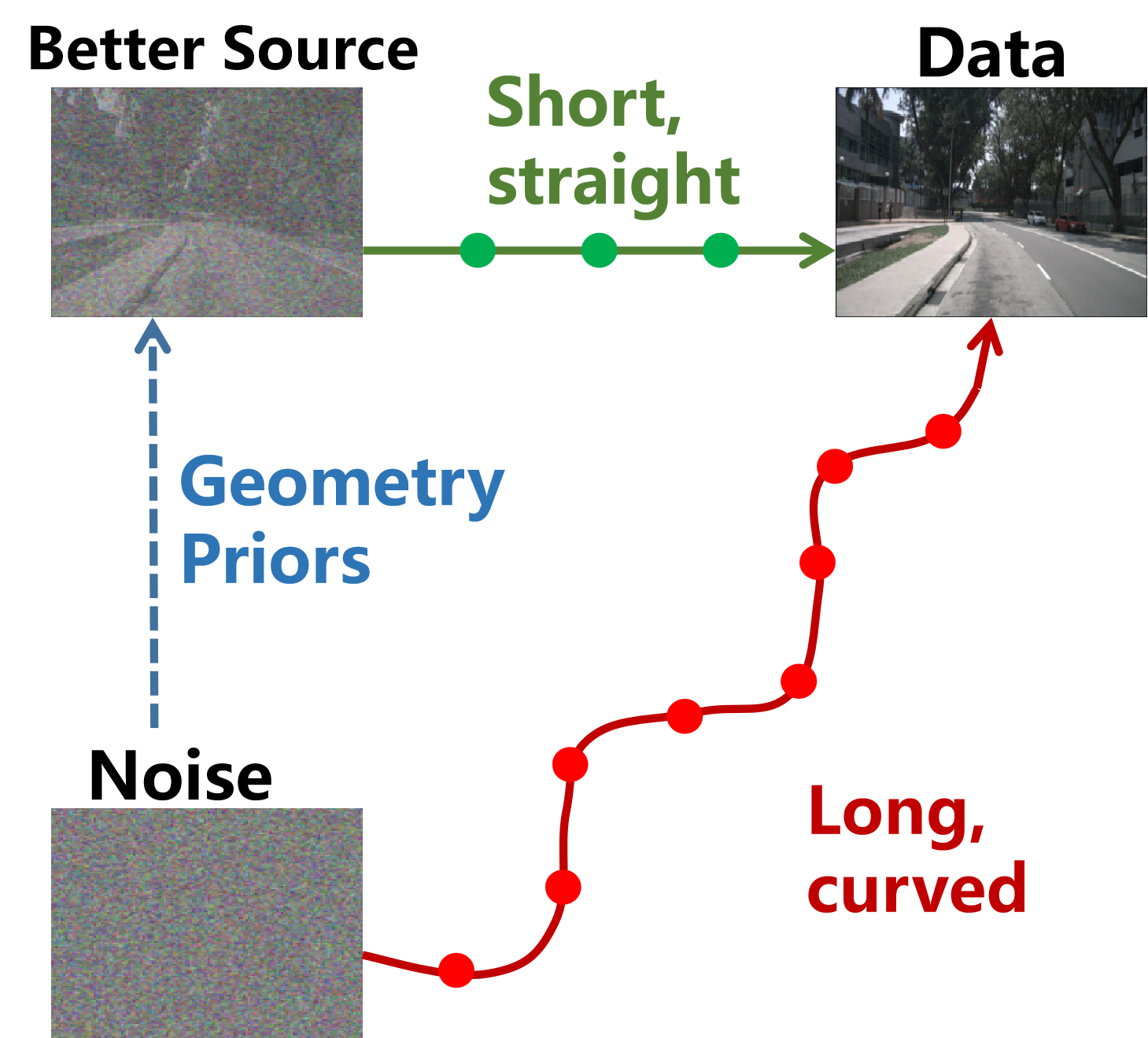}
            \caption{\textbf{Better Starting Point}}
            \label{fig:teaser_principle}
        \end{subfigure}
        
        % \vspace{3mm} % 上下两图之间的间距，根据需要微调
        
        % --- Subfigure (b): Bottom Left ---
        \begin{subfigure}[b]{\linewidth}
            \centering
            % [TODO] 替换为效率曲线 (FVD vs Steps)
            % 注意：因为只有40%宽，这个图最好做成正方形或略高的长方形，不要太扁
            \includegraphics[width=\linewidth, height=3.4cm]{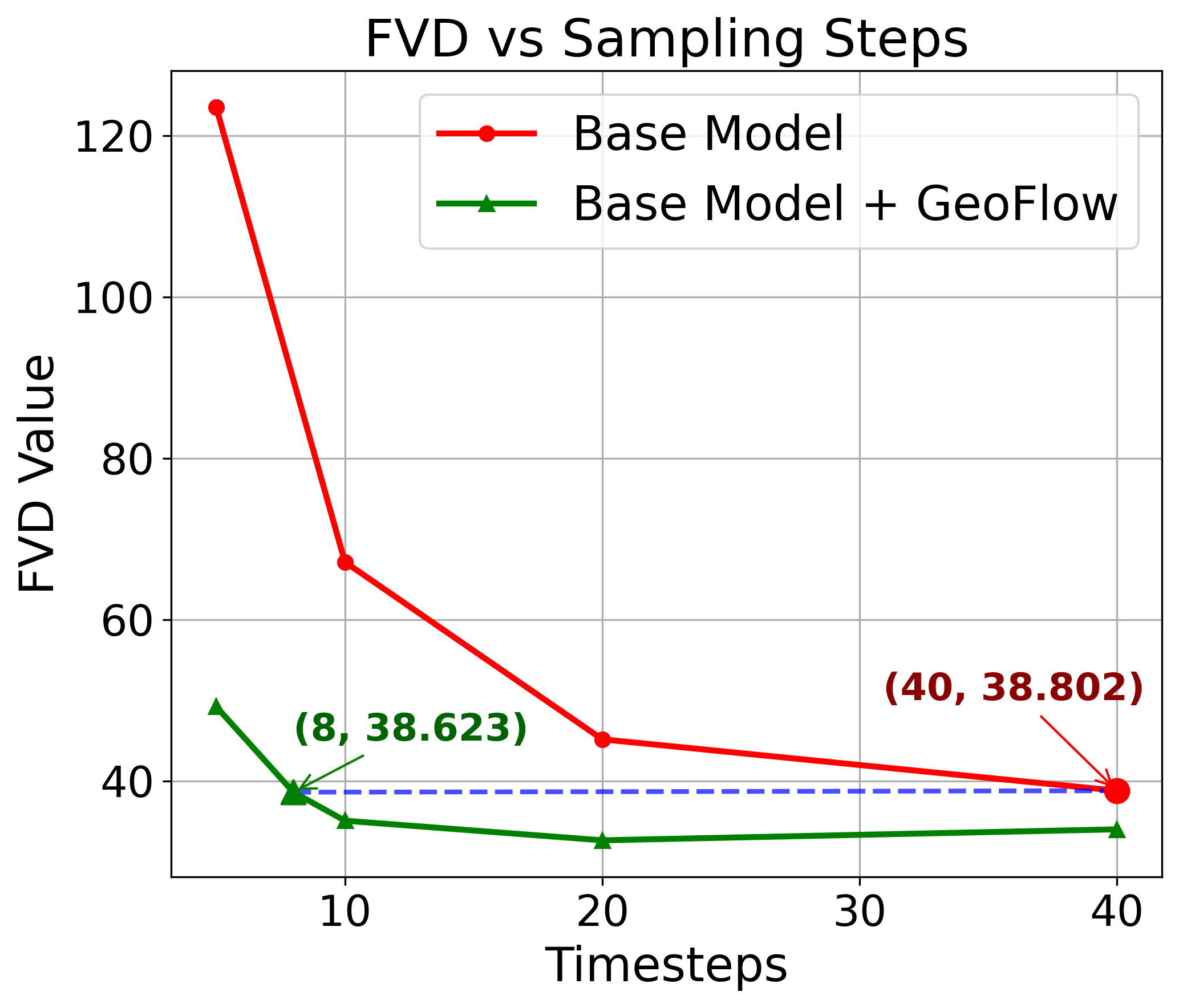}
            \caption{\textbf{Efficiency Analysis}}
            \label{fig:trade_off_curve}
        \end{subfigure}
    \end{minipage}
    \hfill % 左右列之间的弹性间距
    % ==========================================
    % 右侧列 (Right Column): 占 ~58% 宽度 (留2%给间隙)
    % 包含 (c) 视觉对比 (Visual Comparison)
    % ==========================================
    \begin{minipage}[b]{0.67\linewidth}
        % --- Subfigure (c): Right (Full Height) ---
        \begin{subfigure}[b]{\linewidth}
            \centering
            % [TODO] 替换为视觉对比大图
            % 高度建议 = 左边两图高度之和 + 间距 (约 3.8+3.8+0.3 = 7.9cm)
            % 这样底部能对齐，非常整齐
            \includegraphics[width=1.05\linewidth, height=7.5cm]{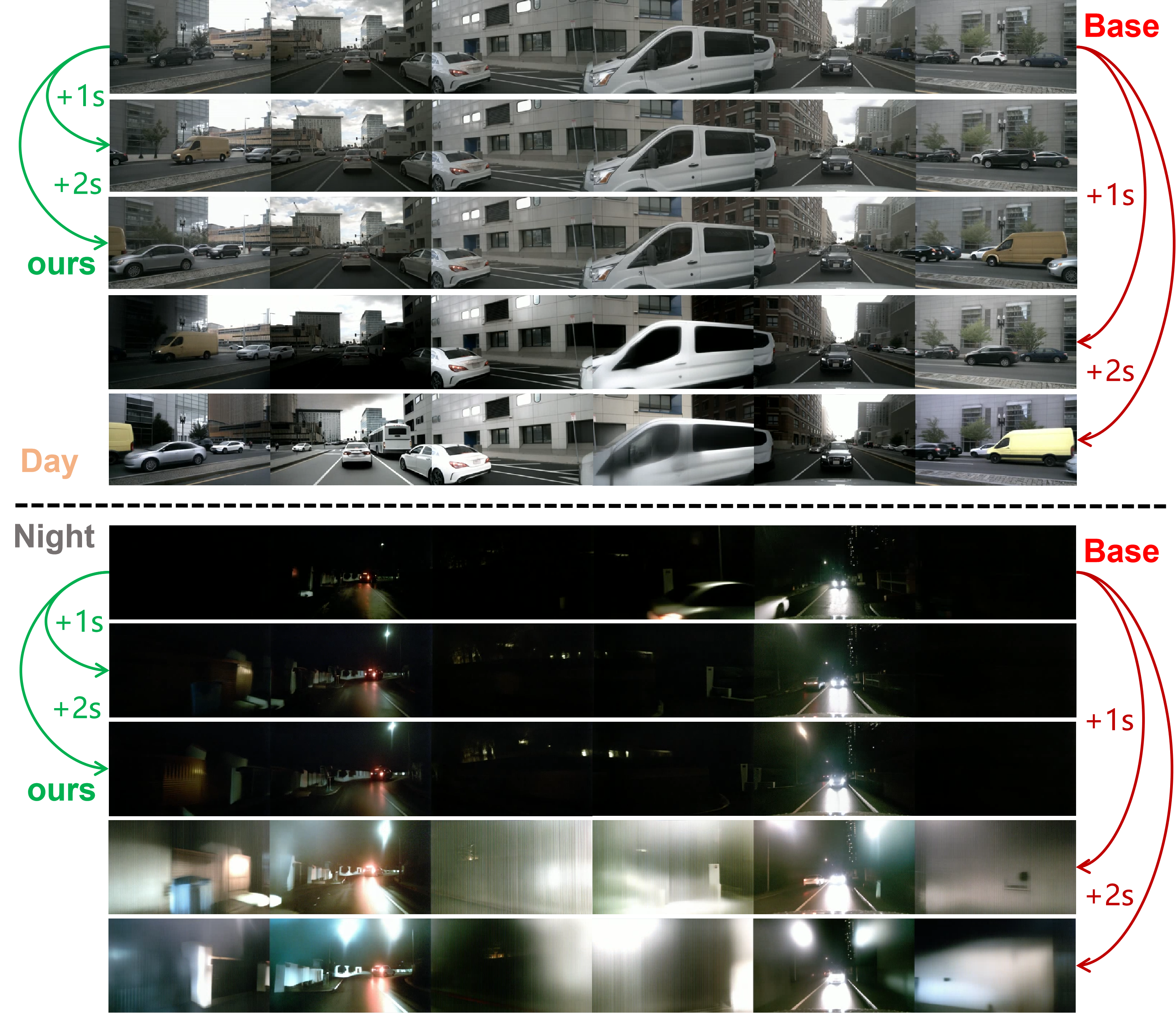}
            \caption{\textbf{Stabilized 5-step Generation}}
            \label{fig:teaser_visual}
        \end{subfigure}
    \end{minipage}

    % ==========================================
    % 总图注 (Main Caption)
    % ==========================================
    \caption{\textbf{Efficient driving video generation with structure-aware priors.} 
    \textbf{(a)} Instead of sampling from noise, we construct a structure-aware prior distribution as a better starting point. 
    \textbf{(b)} This strategy leads to rapid convergence, achieving SOTA quality in just 8 steps (\textbf{\textcolor{mplgreen}{Green line}}) compared to the baseline (\textbf{\textcolor{mplred}{Red line}}). 
    \textbf{(c)} Visual comparisons in 5 steps. 
    % While the baseline (\textbf{\textcolor{mplred}{Red}}) can achieve SOTA performance in 40 steps, it suffers from flickering and ghosting in few steps; 
    Our method (\textbf{\textcolor{mplgreen}{Green}}) significantly improves structural stability and temporal consistency in few-step driving video generation.}
    \label{fig:teaser}
    % \vspace{-0.85cm}
\end{figure*}

% \vspace{-0.5cm}
\section{Introduction}
\label{sec:intro}
% This document serves as an example submission to ECCV \ECCVyear{}.
% It illustrates the format authors must follow when submitting a paper. 
% At the same time, it gives details on various aspects of paper submission, including preservation of anonymity and how to deal with dual submissions.
% We advise authors to read this document carefully.

% The document is based on Springer LNCS instructions as well as on ECCV policies, as established over the years.
The rapid evolution of autonomous driving (AD) systems is increasingly fueled by the availability of massive, diverse, and high-quality driving data. However, acquiring such data from the real world is limited by high costs and safety risks, particularly for scarce corner cases such as extreme weather conditions and hazardous traffic scenarios. To overcome these limitations, generative models\cite{ge2025unraveling,deng2025gaussiandwm,gao2023magicdrive,gao2025magicdrive,wang2024drivedreamer} have emerged as a promising paradigm for synthesizing high-fidelity driving data, serving as a scalable engine for both data augmentation and closed-loop simulation.

Among existing approaches, diffusion probabilistic models and flow matching frameworks have recently established themselves as the state-of-the-art. These generative models offer superior visual fidelity, diversity, and precise controllability over 3D geometry and scene layout. 
Representative works, such as MagicDrive \cite{gao2023magicdrive,gao2025magicdrive} and DriveDreamer \cite{wang2024drivedreamer,zhao2025drivedreamer4d}, have demonstrated remarkable capabilities in generating photorealistic, multi-view driving videos conditioned on rigorous control signals. 
Beyond achieving controllable data generation, these pioneering works demonstrate a potential for establishing generative closed-loop simulation environments\cite{yang2025drivearena,yan2025drivingsphere}, serving as interactive platforms for the closed-loop evaluation and training of autonomous driving algorithms.

However, this high fidelity comes at a steep computational cost. These methods typically rely on iteratively solving Ordinary Differential Equations\cite{song2020denoising,lipman2022flow,liu2022rectifiedflow} (ODEs) or Stochastic Differential Equations\cite{ho2020denoising} (SDEs) to map a standard Gaussian distribution to the complex video data distribution (\textcolor{mplred}{Red} curve in Fig.~\ref{fig:teaser_principle}). 
This process requires extensive sampling steps (often dozens or hundreds) during inference, resulting in significant latency and computational overhead. This inefficiency significantly inflates the cost of data production and severely compromising the interaction frequency of generative simulation pipelines.

\begin{figure}[t]
    % \vspace{-8mm}
    \centering
    \begin{subfigure}[t]{0.44\linewidth}
        \centering
        % [TODO] replace with your vanilla baseline visualization
        \includegraphics[height=2.8cm]{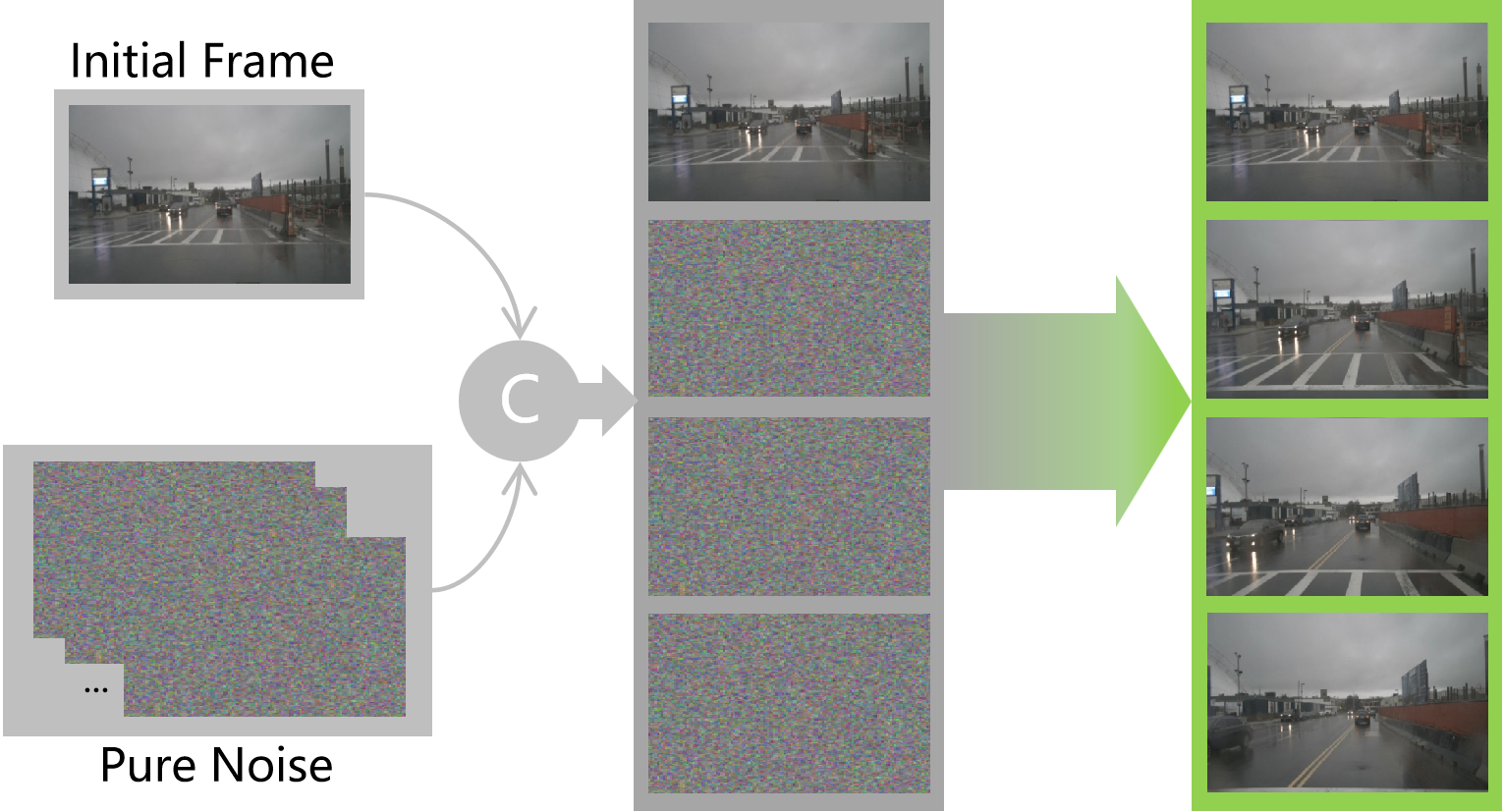}
        \caption{Vanilla}
        \label{fig:vanilla_vis}
    \end{subfigure}
    \hfill
    \begin{subfigure}[t]{0.54\linewidth}
        \centering
        % [TODO] replace with your GeoFlow visualization
        \includegraphics[height=2.8cm]{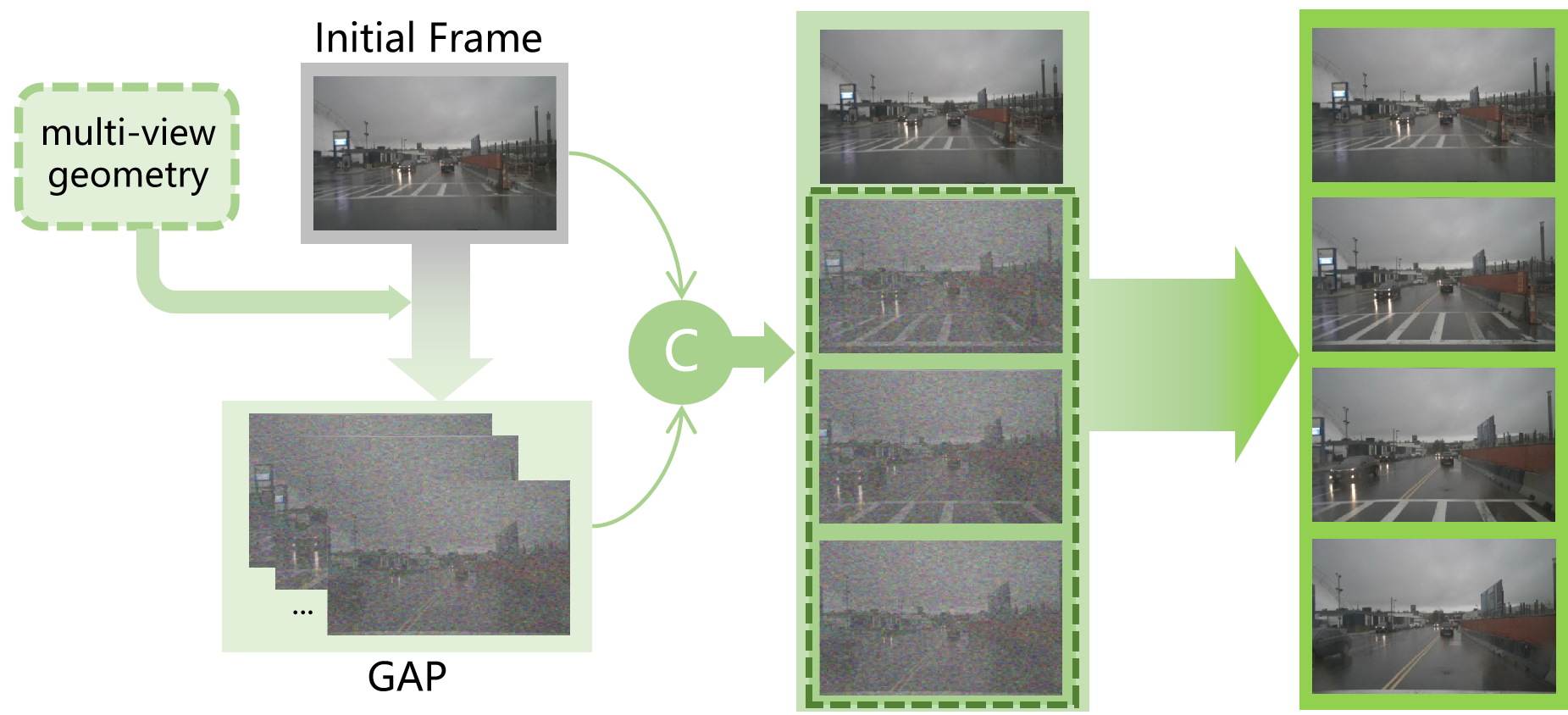}
        \caption{GeoFlow}
        \label{fig:geoflow_vis}
    \end{subfigure}
    \caption{Comparison between Vanilla Driving Video Generation and \textbf{GeoFlow}. The former uses standard Gaussian as source distribution, while GeoFlow leverages multi-view geometry to build an informative source distribution 
    % \textcolor{orange}{in latents for driving scenes}.
    in latent space.
    }
    \label{fig:vanilla_vs_geoflow}
    % \vspace{-8mm}
\end{figure}

We identify that a fundamental source of this inefficiency lies in the ubiquitous assumption that the source distribution is typically \textit{standard Gaussian noise} (Fig.~\ref{fig:vanilla_vis}).
Under this assumption, the model is forced to reconstruct the entire scene, including road, buildings, and vegetation, from pure noise for every single frame.
This formulation ignores the strong spatiotemporal correlations between frames in driving videos, 
% where the background geometry is largely persistent and predictable. 
consequently, not only leading to significant computational redundancy (generating existed contents in history frames) but also increasing the difficulty of maintaining temporal consistency, as initiating consecutive frames from independent noise often results in texture flickering and geometric drift (\textcolor{mplred}{Base} in Fig.~\ref{fig:teaser_visual}), especially in few-step generation.

In the context of autonomous driving, we argue that the source distribution assumption should be revisited. 
Unlike open-domain videos, driving scenarios exhibit remarkably strong \textit{spatiotemporal coherence} and \textit{predictability}. 
The visual evolution of the scene is strictly governed by multi-view geometry and the ego-vehicle's motion: static background elements shift continuously according to the camera extrinsics, while dynamic agents follow trajectories in given conditions.
This implies that given a reference frame and future condition signals, the content of the subsequent frames is largely predictable rather than random. 
Motivated by this insight, we propose a novel perspective: \textit{Why not start the generation from a coarse prediction of the future, rather than from pure noise?}
% Crucially, our prior is \emph{not} a naive "use-the-previous-frame" initialization.
% Instead, we explicitly model the view transformation by unprojecting the reference features with metric depth, transforming them via ego-motion in $SE(3)$, and re-projecting them to the target view while accounting for visibility/occlusion.
% This produces an informative source distribution that is spatially aligned with the target manifold.
% Geometrically, this operation significantly shortens the optimal transport path between the source and target distributions, thereby "straightening" the probability flow and facilitating high-quality generation with fewer sampling steps.

A straightforward solution is to initialize the generation directly from the given reference frame or history frame\cite{liu2025generative}. However, this cannot explicitly account for the multi-frame dynamics induced by ego-motion and dynamic agents, making it difficult to model highly dynamic driving scenes.
We propose to instead leverage metric depth estimation\cite{keetha2025mapanything,yang2024depth, lin2025depth} to warp the reference frame to future poses, producing a geometry-aligned coarse prediction. Nevertheless, the warped result alone is still insufficient to serve as a good source distribution: due to depth estimation inaccuracies and the dynamic nature of the environment, it inevitably contains artifacts.
% Strictly following such a flawed prior would mislead the model, forcing it to generate unrealistic textures to align with geometric errors.
To address this, we propose a Spatially-Adaptive Noise Injection strategy to construct a robust \textbf{Geometry-Aligned Prior Distribution}, where we derive a continuous pixel-wise reliability mask based on geometric consistency and semantic priors to handle artifacts without extra learnable parameters.
% This strategy allows us to adaptively interpolate between the informative warped features and standard Gaussian noise, dynamically modulating the noise intensity according to local uncertainty.
% This formulation naturally guides the model to preserve valid historical structures while focusing its generative capacity on correcting artifacts and completing occluded regions.
As shown in Fig.~\ref{fig:teaser}, this strategy successfully shortens the transport path from the source distribution and the target distribution, enabling high quality driving video generation within several sampling steps.

Our contributions are summarized as follows:
\begin{itemize}
    \item 
    We propose \textbf{GeoFlow}, an efficient driving video generation framework that generates future frames starting from reference frame information rather than pure noise, fundamentally straightening and shortening the transport path.
    \item 
    We construct a \textbf{Geometry-Aligned Prior} distribution that leverages the inherent geometric consistency of driving videos, where \textbf{Spatially-Adaptive Noise Injection} is integrated to effectively mitigate artifacts.
    \item 
    Extensive experiments demonstrate that GeoFlow can significantly improve the few-step generation quality at a low adapting cost.
\end{itemize}
% \begin{itemize}
%     \item 
%     We propose \textbf{GeoFlow}, an efficient driving video generation framework that generates future frames starting from reference frame infomation, fundamentally straightening and shortening the transport path.
%     \item 
%     We construct a \textbf{Geometry-Aligned Prior} distribution that leverages the inherent geometric consistency of driving videos to bridge the gap between the source and target distributions, and further employ \textbf{Spatially-Adaptive Noise Injection} to effectively mitigate artifacts caused by occlusions, dynamic agents, and depth uncertainty.
%     \item 
%     Extensive experiments demonstrate that GeoFlow can achieve remarkable efficiency of both training and inference, significantly improving few-step generation quality at extremely low adapting cost.
% \end{itemize}

% =============================================================================
% Section: Related Work
% =============================================================================
\section{Related Work}
\label{sec:related_work}

% -----------------------------------------------------------------------------
% 2.1 Video Generation for Autonomous Driving
% -----------------------------------------------------------------------------

\subsection{Video Generation for Autonomous Driving}
Data synthesis and closed-loop simulation are of great importance for autonomous driving systems,
% 传统重建方法在自动驾驶数据合成等任务上存在保真度和多样性的限制
while traditional reconstruction methods\cite{gao2026rad, li2026geosvr, gu2026sparsesurf, zhang2026eve3d} have limitations in fidelity and diversity.
Generative models\cite{goodfellow2020generative,ho2020denoising,lipman2022flow,liu2022rectifiedflow} thus have become indispensable to the autonomous driving lifecycle.
% While early approaches relied on GANs \cite{goodfellow2020generative}, recent advancements are predominantly driven by Diffusion Probabilistic Models (DPMs) \cite{ho2020denoising} and Flow Matching frameworks \cite{lipman2022flow,liu2022rectifiedflow} due to their superior stability and synthesis quality.
DriveDreamer \cite{wang2024drivedreamer} pioneered the incorporation of structural constraints, achieving controllable video synthesis conditioned on Bird’s-Eye-View (BEV) maps and 3D bounding boxes. 
MagicDrive\cite{gao2023magicdrive} further extended these capabilities to ensure strict multi-view geometric consistency. Subsequent research\cite{gao2025magicdrive} has continued to push the boundaries of controllability, spatial resolution, and temporal duration, leading to significant leaps in the fidelity and realism of synthesized driving scenarios.
Fueled by this progress, the scope of video generation has expanded beyond dataset creation to more tasks such as neural scene reconstruction\cite{zhao2025drivedreamer4d,ni2025recondreamer,zhao2025recondreamer++} and closed-loop simulation\cite{yang2025drivearena,yan2025drivingsphere}.
However, despite these successes, a fundamental challenge persists: these methods inherently require a large number of denoising steps (sampling iterations) to obtain high-quality results. 
This requirement creates a conflict between inference efficiency and generation quality, 
% creating a prohibitive computational burden for large-scale data production. More critically, the high latency drastically impedes the interaction frequency required for responsive generative environments, 
thereby hindering the efficient deployment of applications like generative closed-loop simulation.
% To address this efficiency bottleneck, we propose GeoFlow, a novel framework that significantly reduces the number of denoising steps required for high-fidelity synthesis, achieving efficient realistic driving video generation.

% -----------------------------------------------------------------------------
% 2.2 Efficient Video Generation
% -----------------------------------------------------------------------------
% 这段太像causvid！
% \vspace{-2mm}
\subsection{Efficient Video Generation}
% \vspace{-1mm}

To mitigate the high inference latency of diffusion models, existing acceleration strategies primarily focus on designing advanced samplers \cite{song2020denoising, lu2022dpm, zhao2023unipc} to optimize the numerical solution of ODE, or employing model distillation\cite{salimans2022progressive, song2023consistency, liu2022rectifiedflow, yin2024one} to compress the multi-step generative trajectory, including recent advances in the video domain \cite{mao2025osv, wang2023videolcm, zhang2024sf, yin2025slow, huang2025self}. Despite enabling rapid synthesis, distillation-based methods typically incur expensive training costs. And these approaches retain the standard Gaussian source assumption, essentially ignoring the strong spatiotemporal and geometric correlations inherent in driving scenarios. In contrast, GeoFlow tackles efficiency from the perspective of source distribution construction. By establishing a geometry-aligned prior, our method naturally shortens and straightens the generative trajectory, which is orthogonal to existing frameworks.

% Despite their effectiveness, these methods typically require expensive distillation costs, more critically, they predominantly focus on optimizing solving path or distilling teacher models while retaining the standard Gaussian source assumption. This ignores the strong geometric correlations in driving scenarios. 
% In contrast, GeoFlow constructs a geometry-aligned prior distribution, naturally shortening and straightening the generative trajectory, achieving acceleration without complex distillation pipelines. 
% This also suggests that GeoFlow approaches efficiency from the perspective of building a better source distribution, which is orthogonal to existing methods and can be seamlessly integrated into their frameworks.

% -----------------------------------------------------------------------------
% 2.3 Geometry-Aware Synthesis
% -----------------------------------------------------------------------------
% \vspace{-2mm}
% \subsection{Mapping between Arbitrary Distributions}
\subsection{Beyond Standard Gaussian Source Distribution 
% \textcolor{orange}{(Source Distribution in Flow-based Generative Models ?)}
}
% \vspace{-1mm}
To break free from the standard Gaussian source assumption for specific tasks, Diffusion Bridges\cite{zhou2023denoising} extend the standard diffusion framework to learn the transport between two arbitrary probability distributions. 
These methods have achieved superior performance in specific data-to-data translation tasks, such as image super-resolution\cite{yue2023resshift}, in-painting\cite{peng2025stabilizing,han2025asyncdsb}, and restoration\cite{wang2025residual,yue2025enhanced}.
% by leveraging coupled data pairs from two distributions. 
Similarly, Flow Matching (FM)\cite{lipman2022flow,liu2022rectifiedflow} is theoretically grounded in regressing vector fields defined by the interpolation between sample pairs, naturally supporting mappings between any two arbitrary distributions. 
While FM has been successfully applied to data-to-data translation tasks\cite{martin2024pnp,cohen2025efficient,qin2025reversing,lugmayr2020srflow,yun2025flowhigh,wang2022diverse}, mainstream research in video generation predominantly retains the standard Gaussian source assumption to ensure training stability and generation diversity.

In the context of video synthesis, Video Bi-flow\cite{liu2025generative} first investigates this flexibility by using the previous frame with noise injection as the starting point of flow matching. Their experiments demonstrate that evolving the flow from the previous frame yields a shorter transport distance and higher sampling efficiency compared to standard methods that treat the previous frame merely as a condition.
GeoFlow shares a similar insight of starting from a better source distribution, 
yet it fundamentally diverges in how this distribution is constructed. 
We explicitly model the transition between frames using physical priors to construct the source distribution, which makes it geometrically aligned with the target manifold.
% rather than directly mapping from history frame. 
% While Video Bi-flow learns the mapping between consecutive frames, with a simple noise injection strategy to mitigate error accumulation, 
% Our GeoFlow explicitly models the viewpoint transition between frames leveraging geometric priors, and fully exploits the rich condition signals inherent in controllable driving generation (e.g., object and map layout) to perform spatially-adaptive noise injection. This synergy allows us to construct a Geometry-Aligned Prior Distribution, which serves as a physically grounded starting point that is significantly closer to the target data manifold.

% -----------------------------------------------------------------------------
% 2.4 Geometry-Constrained Video Generation
% -----------------------------------------------------------------------------
% \vspace{-3mm}
\subsection{Geometry-Constrained Video Generation}
% \vspace{-1mm}
% 在视频生成任务中引入几何先验是近期研究的热点。CamCtrl将渲染视频与扩散模型的噪声拼接起来输入视频生成模型，以此作为条件信息指导生成，显著提升了视频生成的可控性与一致性。GEN3C[28]将上述的思想进一步扩展，将三维结构作为视频生成的显式缓存，让长视频生成也能依靠显式的三维结构来提升一致性。尽管这些方法都利用了几何先验来指导视频生成，但与我们的方法有以下根本的不同：
% 1. 这些工作的动机一般是通过几何信息来实现精准的位姿控制，或提高视频生成的一致性。而GeoFlow的动机是利用几何先验构造更近的源分布来提高生成效率。
% 2. 这些工作在方法上通常讲几何信息作为condition来注入生成过程，这意味着模型在推理过程中仍然要经历从纯噪声到目标数据的完整路径。而GeoFlow直接从更近的起点出发，transport path更短更直。
Introducing geometric priors into video generation tasks has recently been an active area of research. 
Existing works\cite{ren2025gen3c, he2024cameractrl, hou2024training, yu2024viewcrafter, zhang2025i2v3d, chen2025geodrive} typically integrate 3D geometric information\cite{kerbl20233d, wang2025vggt, zhang2024cor, li2024dngaussian, li2026dngaussian++} as conditioning signals to guide the generation process , achieving precise camera control and enhancing the temporal coherence of synthesized videos. 
% Although GeoFlow and these methods share the similar idea of leveraging geometric priors, they fundamentally differ from motivation and methodology. 
GeoFlow fundamentally differs from these works in both rationale and methodology. 
The primary objective of these prior works is to enforce controllability and consistency through explicit 3D conditioning, meaning the generative model still traverses the complete trajectory from pure Gaussian noise to the target data manifold during inference. In contrast, GeoFlow is designed from the perspective of generation efficiency. Rather than conditioning, we construct a geometrically aligned and much closer source distribution, which significantly shortens and straightens the transport path, thus improving sampling efficiency.

\section{Method}
\label{sec:method}

% -----------------------------------------------------------------------------
% 3.1 Preliminaries: 
% -----------------------------------------------------------------------------
% \subsection{Preliminaries}
% \label{subsec:preliminaries}

% \noindent \textbf{Flow Matching.}
% We formulate the video generation process using the Flow Matching framework, which models the continuous dynamics of data generation via an Ordinary Differential Equation (ODE). Let $x_0 \sim p_0(x)$ denote the source distribution and $x_1 \sim p_{data}(x)$ denote the target data distribution. The probability flow ODE is defined as:
% \begin{equation}
%     \frac{d x_t}{dt} = v_t(x_t, t), \quad t \in [0, 1],
% \end{equation}
% where $v_t$ is a time-dependent vector field that transports samples from $p_0$ to $p_{data}$.
% During training, Flow Matching defines the intermediate state $x_t$ by linear intepolation between data pairs sampled from the two distribution:
% \begin{equation}
%     x_t = (1 - t)x_0 + t x_1.
% \end{equation}
% The model learns the velocity field $v_\theta$ between source and target distribution by optimizing this regression objective:
% \begin{equation}
%     \mathcal{L}_{FM} = \mathbb{E}_{t, x_0, x_1} \left[ || v_\theta(x_t, t) - (x_1 - x_0) ||^2 \right].
% \end{equation}
% \vspace{-2mm}
\subsection{Problem Formulation}
\label{subsec:formulation}
% \vspace{-1mm}
\noindent \textbf{Task Definition.}
Given a past video sequence (or a single reference frame $\mathbf{I}_{ref}$), the goal is to generate the future video chunk $\mathbf{V} \in \mathbb{R}^{L \times N \times H \times W}$ that strictly follows a set of control signals $\mathcal{C}$.
We formulate this as learning a conditional distribution $p(\mathbf{V} | \mathbf{I}_{ref}, \mathcal{C})$.

\noindent \textbf{Control Conditions.}
Following standard practices in driving world modeling~\cite{gao2023magicdrive,wang2024drivedreamer}, the condition set $\mathcal{C}$ encapsulates the geometric and semantic layout of the driving scene: it includes (i) \textbf{ego-motion and camera parameters} $\mathcal{C}_{cam}$, i.e., the camera extrinsics $\mathbf{T} \in SE(3)$ and intrinsics $\mathbf{K}$ for each frame, defining the 3D-to-2D projection rules; (ii) the \textbf{3D object layout} $\mathcal{C}_{obj}$, represented as a set of 3D bounding boxes $\{ (b_k, c_k) \}$ that describe the position and category of dynamic agents (e.g., vehicles, pedestrians) in the future scene; and (iii) a \textbf{road map} $\mathcal{C}_{map}$, such as an HD-Map or BEV layout providing road boundaries and lane information.
% \textit{Crucially, these conditions are standard inputs required for controllable simulation. Our method leverages $\mathcal{C}_{cam}$ and $\mathcal{C}_{obj}$ to construct the geometric prior without requiring any additional user annotations or external data.}

% \noindent \textbf{Flow Matching Backbone.}
% We employ a latent flow matching model parameterized to predict the clean data $x_1$ (latent representation of $\mathbf{V}$).
% The generative process is governed by the ODE: $d x_t / dt = v_t(x_t, t, \mathcal{C})$. The model $F_\theta(x_t, t, \mathcal{C})$ is trained to minimize the flow matching objective:
% \begin{equation}
%     \mathcal{L} = \mathbb{E}_{t, x_1, x_0, \mathcal{C}} \left[ || F_\theta(x_t, t, \mathcal{C}) - x_1 ||^2 \right],
% \end{equation}
% where $x_t = (1-t)x_0 + t x_1$.
\noindent \textbf{Flow Matching Backbone.}
We employ a latent flow matching model parameterized to estimate the vector field $v_t$ towards the clean data $x_1$ (latent representation of $V$).
The generative process is governed by the ODE: $d x_t / dt = v_t(x_t, t, \mathcal{C})$. The model $F_\theta(x_t, t, \mathcal{C})$ is trained to minimize the flow matching objective:
\begin{equation}
    \mathcal{L} = \mathbb{E}_{t, x_1, x_0, \mathcal{C}} \left[ || F_\theta(x_t, t, \mathcal{C}) - v_t ||^2 \right],
\end{equation}
where $x_t = (1-t)x_0 + t x_1$ and $v_t = x_1 - x_0$.

% Standard methods assume $x_0 \sim \mathcal{N}(0, \mathbf{I})$. Our contribution lies in redefining $x_0$ by explicitly utilizing $\mathbf{I}_{ref}$ and $\mathcal{C}$.

\begin{figure}[tb]
  \centering
  \includegraphics[height=5.0cm, width=1.05\linewidth]{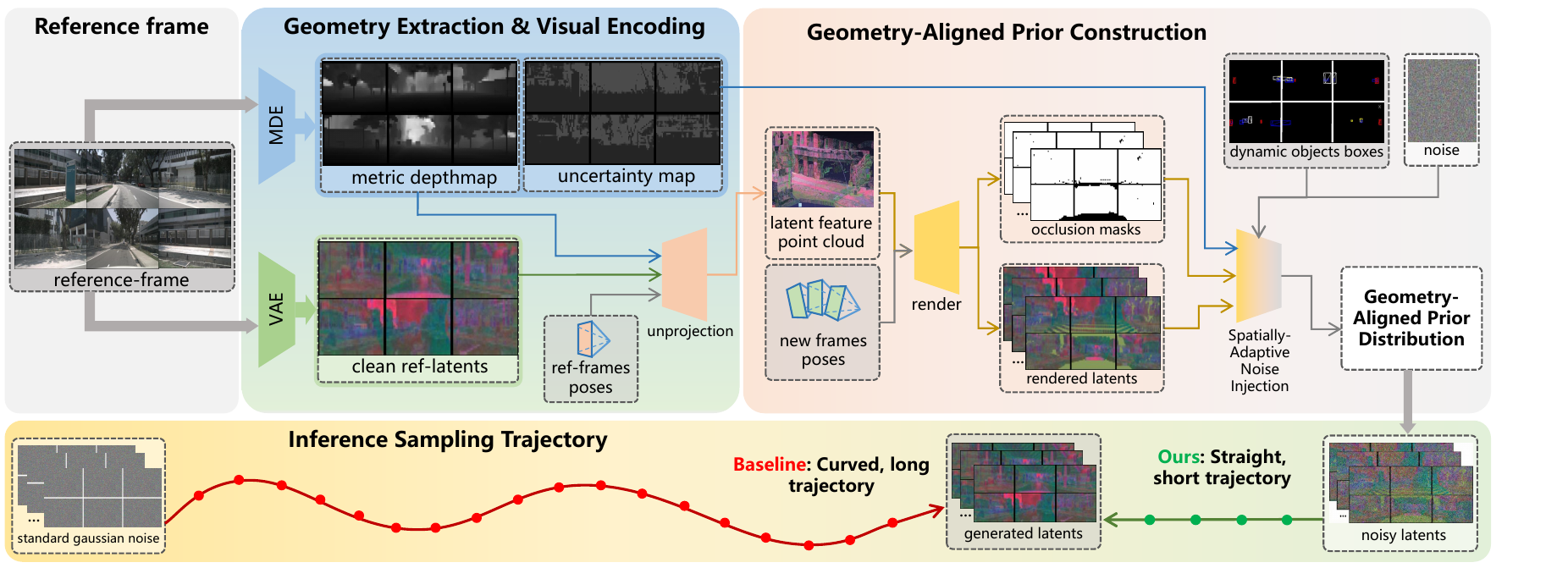} % 建议用 width=\linewidth 自适应宽度
  \caption{\textbf{Schematic of the proposed GeoFlow framework.} 
  % Instead of initializing from an uninformative noise, 
  % Our method leverages multi-view geometry to bridge the gap between the source and target manifolds.
  % The pipeline projects visual features into a latent point cloud, which is rendered to the target view to form a geometric prior.
  To bridge the gap between the source and target manifolds, we leverage multi-view geometry to construct a \textbf{Geometry-Aligned Prior Distribution}, where an Spatially-Adaptive Noise Injection strategy is introduced to reduce error accumulation due to depth and rendering artifacts.
  \textbf{Bottom Panel:} As visualized in the sampling trajectory comparison, our initialization significantly \textit{shortens and straighten} the flow, 
  % yielding a nearly straight generation path (green) compared to the highly curved trajectory of the standard Gaussian baseline (red), 
  thereby enabling high-fidelity synthesis within several inference steps.
  }
  \label{fig:framework}
  % \vspace{-5mm}
\end{figure}

% -----------------------------------------------------------------------------
% 3.3 GeoFlow Framework
% -----------------------------------------------------------------------------
\subsection{GeoFlow Framework}
\label{subsec:geoflow_framework}

Standard driving video generation models typically initialize from pure Gaussian noise ($x_0 \sim \mathcal{N}(0, \mathbf{I})$), forcing the model to construct complex driving scenes entirely from randomness. This leads to highly non-linear generation trajectories and requires numerous integration steps to solve, which fundamentally limits the generation efficiency. 
We address this problem by introducing the \textbf{GeoFlow} framework. 
Its core principle is to replace the standard Gaussian noise with a \textbf{Geometry-Aligned Prior Distribution}. 
This makes the source distribution geometrically aligned with the target, thus significantly shortening and straightening the generation path, enabling efficient high-fidelity driving video synthesis with substantially fewer sampling steps (Fig.~\ref{fig:framework}).

The following sections detail the construction process of this Geometry-Aligned Prior Distribution, which is primarily divided into two subsections: Latent Geometry Extraction (Section~\ref{subsec:geometry}) and Geometry-Aligned Prior Construction (Section~\ref{subsec:prior_construction}). Finally, we introduce the training and inference pipelines of GeoFlow (Section~\ref{subsec:training}).

% \vspace{-3mm}
% \subsection{Latent Warping}
% However, driving videos exhibit strong spatiotemporal coherence: the visual evolution of static backgrounds is strictly governed by the ego-vehicle's motion, making future scene layouts highly predictable given historical frames.
% \vspace{-2mm}
\subsection{Latent Geometry Extraction}
\label{subsec:geometry}
% \vspace{-1mm}
Driving scenarios are characterized by strong spatiotemporal coherence. 
Unlike general open-domain videos, the visual evolution of a driving scene is heavily correlated with the ego-vehicle's control signals. 
Specifically, the static scene structures (e.g., roads, buildings, vegetation), which typically occupy the majority of the field of view, shift continuously in accordance with the camera's ego-motion. 
This observation implies that a significant portion of the future frame content is not random but can be explicitly derived from historical frames and given conditions. 

Motivated by this insight, we propose \textbf{explicitly} modeling the geometric relationships between sequential frames to reduce the computational redundancy in video generation.
Specifically, we leverage the known future ego-motion to warp the reference frame to expected future locations. 
This allows the model to start directly from a \textbf{coarse prediction} based on historical context, eliminating the need to redundantly regenerate existing content.

% \noindent \textbf{Process.}
Instead of operating in RGB space, we choose the latent space of the VAE for efficiency.
First, we estimate the metric depth $\mathbf{D}_{ref}$ of the reference frame, which also produces the corresponding uncertainty map $\mathbf{M}_{unc}$ to describe the accuracy of depth estimation. And we encode reference frame to get latent representation $\mathbf{Z}_{ref} = \mathcal{E}(\mathbf{I}_{ref})$, then unproject the latent features $\mathbf{Z}_{ref}$ into a 3D feature point cloud $\mathcal{P}_{ref}$ using estimated metric depth and camera parameters, as shown in Fig.~\ref{fig:framework}.
Given the relative pose transformation $\mathbf{T}_{rel}$ derived from the ego-vehicle's control commands or trajectory planning, we transform the point cloud to the target coordinate system:
\begin{equation}
    \mathcal{P}_{target} = \mathbf{T}_{rel} \cdot \mathcal{P}_{ref}.
\end{equation}
We then render $\mathcal{P}_{target}$ using splatting algorithm to obtain the warped latent map $\mathbf{Z}_{warp}$. 
This process creates a geometrically aligned canvas for the future frame. Detailed computational procedures are provided in the supplementary materials.
% 详细的计算过程见附录

% 在之前的 "Latent Point Cloud Rendering" 部分，插入关于 Feature Splatting 的细节
% \vspace{0.1cm}
\noindent \textbf{Feature Splatting Mechanism.}
Projecting 3D points to the 2D image plane may result in collisions (multiple points mapping to the same pixel). To address collisions and keep the input shape aligned with base model, we adopt a Z-buffer based feature splatting strategy.

Let $\mathcal{P}_{target} = \{ (\mathbf{p}_k, \mathbf{f}_k) \}$ denote the set of transformed 3D points $\mathbf{p}_k$ and their associated feature vectors $\mathbf{f}_k$. For a target pixel coordinate $\mathbf{u} = (u, v)$, we identify the subset of points $\mathcal{N}(\mathbf{u})$ that project onto this pixel neighborhood. The value of the warped latent map $\mathbf{Z}_{warp}(\mathbf{u})$ is determined by the point closest to the camera:
\begin{equation}
    k^* = \arg\min_{k \in \mathcal{N}(\mathbf{u})} d_z(\mathbf{p}_k),
\end{equation}
% \vspace{-1mm}
\begin{equation}
    \mathbf{Z}_{warp}(\mathbf{u}) = \mathbf{f}_{k^*},
\end{equation}
where $d_z(\cdot)$ denotes the depth value in the camera coordinate system.
% This "Hard Splatting" approach ensures that foreground features correctly occlude background features in the latent space, preserving sharp boundaries for objects like vehicles and guardrails.

% -----------------------------------------------------------------------------
% 3.3 Uncertainty-Modulated Prior Construction: 噪声注入
% -----------------------------------------------------------------------------
% \vspace{-0.2cm}
\subsection{Geometry-Aligned Prior Construction}
\label{subsec:prior_construction}
% \vspace{-0.1cm}
% \noindent \textbf{Motivation: Determinism vs. Stochasticity.}
Directly utilizing the deterministic warped latents $\mathbf{Z}_{warp}$ as the source distribution $\hat{x}_0$ is suboptimal for two main reasons.
First, Flow Matching models fundamentally require a stochastic source distribution to ensure training stability and generation diversity. Initializing from a deterministic point restricts the model's ability to explore the data manifold, potentially leading to severe error accumulation in inference\cite{liu2025generative}.
% , our ablation study also verified this point (Sec.~\ref{para:noise_injection}).
Second, the geometric prior is inherently imperfect. Due to inevitable inaccuracies in metric depth estimation and rendering artifacts, $\mathbf{Z}_{warp}$ contains structural distortions. 
Strictly following these flawed features would force the model to hallucinate unrealistic textures to match the geometric errors.
Therefore, it is crucial to introduce stochastic perturbations to mitigate artifacts and encourage the generative model to naturally correct and inpaint the warped $\mathbf{Z}_{warp}$.

\noindent \textbf{Limitations of Global Noise.}
A straightforward solution to introduce stochasticity and mitigate artifacts is to inject global uniform noise (i.e., $\hat{x}_0 = (1-\alpha) \mathbf{Z}_{warp} + \alpha \epsilon$). 
However, our experiments suggest that this is a compromise: while it aids convergence in few-step inference, it degrades performance at higher sampling steps compared to the baseline. 
This is because uniform noise indiscriminately corrupts the artifacts and the high-quality regions of the warped prior, wasting the valuable geometric information we extracted.

\noindent \textbf{Spatially-Adaptive Noise Injection.}
To resolve this dilemma, we propose a Spatially-Adaptive Noise Injection strategy. 
Instead of treating all regions equally, we construct a continuous noise injection mask $\mathbf{M}$ to modulate the noise intensity locally. Specifically, low values in the mask identify reliable regions where the extracted geometric prior should be strictly preserved. Conversely, high values signify unreliable areas, such as occlusions or dynamic objects, instructing the model to discard the flawed prior and rely predominantly on standard Gaussian noise for those specific regions.

% \noindent \textbf{Spatially-Adaptive Noise Injection.}
% To resolve this dilemma, we propose a Spatially-Adaptive Noise Injection strategy. Instead of treating all regions equally, we construct a continuous Noise Injection Mask $\mathbf{M} \in [0, 1]$ to modulate the noise intensity locally. 
% Here, $\mathbf{M} \approx 0$ indicates reliable regions where the geometric prior is preserved, while $\mathbf{M} \approx 1$ denotes unreliable regions (e.g., occlusions, dynamic objects) where the distribution is dominated by standard Gaussian noise.

The mask aggregates three sources of uncertainty, each corresponding to a specific geometric error type and thus providing a controllable fallback mechanism. 
The first source is geometric invalidity $\mathbf{M}_{occ}$, derived from the Z-buffer during rendering, where regions that are occluded or out-of-view in the reference frame are assigned high noise intensity. 
The second source $\mathbf{M}_{dyn}$ addresses dynamic objects, as static geometry warping cannot predict object motion; we utilize projected 3D bounding boxes to mask out regions corresponding to dynamic agents, allowing the generative model to redraw their poses. 
The third source is depth uncertainty $\mathbf{M}_{unc}$, where we utilize the uncertainty map from the depth estimator (normalized to $[0, 1]$) to assign corresponding noise levels to regions with ambiguous geometry, such as sky or transparent surfaces.
% More details like the visualization of these masks and warped latents with or without noise injection are shown in supplementary materials.

The final noise injection mask is obtained by taking the maximum operation across these components:
\begin{equation}
    \mathbf{M} = \max(\mathbf{M}_{occ}, \mathbf{M}_{dyn}, \mathbf{M}_{unc}).
\end{equation}

\noindent \textbf{Noise Injection.}
We construct the final source distribution $\hat{x}_0$ via linear interpolation between the warped prior and standard Gaussian noise $\epsilon$, controlled by $\mathbf{M}$:
\begin{equation}
    \hat{x}_0 = (1 - \mathbf{M}) \odot \mathbf{Z}_{warp} + \mathbf{M} \odot \epsilon, \quad \epsilon \sim \mathcal{N}(0, \mathbf{I}).
\end{equation}
% -----------------------------------------------------------------------------
% 3.4 
% -----------------------------------------------------------------------------
% \vspace{-0.6cm}
\subsection{Training and Inference}
\label{subsec:training}
% \vspace{-0.1cm}
\textbf{Chunk Strategy.} 
Our method relies on the geometric overlap between the reference and target views. To ensure valid reprojection, we adopt a short-horizon frame autoregressive strategy, setting the chunk size to $L=6$ which contains 1 reference frame and 5 new generated frames. 
% While this requires multiple autoregressive inference passes to generate long videos (e.g., 4 passes for a 16-frame clip), our method still achieve more than 4x acceleration and improve the interaction frequency up to 2Hz (for a 10 fps video).
% Formally, the complete initial state $\mathbf{X}_0$ fed into the generative model is constructed by concatenating the clean latent representation of the reference frame $\mathbf{Z}_{ref}$ and our constructed future  $\hat{x}_0$ along the sequence (temporal) dimension:
% \begin{equation}
%     \mathbf{X}_0 = [\mathbf{Z}_{ref} ; \hat{x}_0],
% \end{equation}
% where $[\cdot ; \cdot]$ denotes temporal concatenation. This operation yields a complete source state of length $L$, ensuring that the model conditions on the exact historical appearance while seamlessly generating the subsequent frames along the probability flow.

\noindent \textbf{Optimization Objective.}
Following the baseline configuration, we maintain the vector field prediction parameterization where the network $F_\theta$ estimates the optimal transport velocity $v_t$. We fine-tune the baseline model to adapt to the proposed Geometry-Aligned Prior Distribution. The intermediate training states $x_t$ are sampled via linear interpolation along the optimal transport path between our constructed source $\hat{x}_0$ and the ground truth $x_1$:
\begin{equation}
    x_t = (1 - t)\hat{x}_0 + t x_1, \quad t \in [0, 1].
\end{equation}
The training objective is to minimize the prediction error of the vector field:
\begin{equation}
    \mathcal{L} = \mathbb{E}_{t, x_1, \hat{x}_0, \mathcal{C}} \left[ || F_\theta(x_t, t, \mathcal{C}) - v_t ||^2 \right],
\end{equation}
where the analytical velocity is defined as $v_t = x_1 - \hat{x}_0$.

\noindent \textbf{Inference.}
During inference, instead of sampling from a standard Gaussian distribution, we initialize the generative process directly with our Geometry-Aligned Prior Distribution ($x_0 = \hat{x}_0$). 
From this starting point, we employ standard numerical ODE solvers provided by the baseline architecture to integrate the estimated vector field $F_\theta(x_t, t, \mathcal{C})$ over time.

\section{Experiments}
\label{sec:experiments}

\subsection{Experimental Setup}
\label{subsec:setup}
% \vspace{-0.2cm}
\noindent \textbf{Dataset.} 
We conduct our experiments on the NuScenes dataset\cite{caesar2020nuscenes}, a large-scale autonomous driving benchmark consisting of 1000 scenes (700 for training, 150 for validation, and 150 for testing). Each scene spans 20 seconds and captures 360-degree views via six cameras. 
We utilize the training split for fine-tuning baseline models and the validation split for evaluation. The video frames are resized to a resolution of $256 \times 448$ for training and inference.
\noindent \textbf{Baselines.} 
We build our framework upon \textbf{OpenDWM}~\cite{opendwm} codebase, which reproduces state-of-the-art driving video generation methods and open-source the training, inference, and evaluation codes. We compare our approach primarily against the baseline with Gaussian initialization.
Further details, such as a detailed overview of this codebase and the specific selection of the baseline model, are provided in the supplementary materials.
To ensure a fair comparison, both the baseline and our model share the exact same model architecture and control conditions, only differ in source distribution.

\noindent \textbf{Metrics.} 
We assess the generation performance from two perspectives:
(1) \textit{Video Quality:} We report \textbf{FVD} (Fréchet Video Distance) computed on clips of 16 frames to measure the temporal coherence and visual fidelity.
(2) \textit{Image Quality:} We report \textbf{FID} (Fréchet Inception Distance) to evaluate the visual quality of individual frames.
Following UniMLVG\cite{chen2024unimlvg}, we conduct evaluation using 150 scenes from NuScenes validation set, and generate six-view 16-frame videos to calculate those metrics.
Except where noted, each video is generated starting from one reference frame.
% Each video is generated starting from the first frame of the corresponding scene.

\noindent \textbf{Implementation Details.} 
We initialize our model from the pre-trained weights of OpenDWM. We fine-tune the pre-trained model on 2 NVIDIA H100 GPUs to adapt to the proposed Geometry-Aligned Prior Distribution. The learning rate is set to 5e-5. 
During training, we use MapAnything-v1.0\cite{keetha2025mapanything} as metric depth estimator, and we also conduct experiments on other metric depth models during inference.

% -----------------------------------------------------------------------------
% 4.2 Main Results
% -----------------------------------------------------------------------------
% \vspace{-0.5cm}
\subsection{Main Results}
\label{subsec:main_results}
% \vspace{-0.2cm}

\noindent \textbf{Quantitative Comparison.}
Table~\ref{tab:main_results} compares our method with other SOTA approaches with standard Gaussian initialization. Although existing methods achieve high-fidelity synthesis, they usually require many inference steps. 
In contrast, GeoFlow synthesizes driving videos with a lower FVD than the baseline (referred to as OpenDWM) using significantly fewer inference steps. 
This demonstrates that our geometry-aligned prior not only improves the efficiency of the generative process but also naturally enforces better spatiotemporal consistency across frames.

\begin{table}[t]
    \centering
    \label{tab:combined_results}
    \caption{\textbf{Quantitative comparison on NuScenes validation set.} (a) We compare our GeoFlow with state-of-the-art methods. Our \textbf{GeoFlow} achieves SOTA quality with significantly less steps. $^\dagger$ means the metrics are calculated using 3 reference frames. (b) We conduct experiments on base models with different architecture, GeoFlow framework can bring significant improvement on few-step driving video generation.}
    % ================= LEFT BIG TABLE =================
    % \vspace{-2mm}
    % \hspace{-6mm}
    \begin{subtable}[t]{0.52\textwidth}
        \centering
        \setlength{\tabcolsep}{3.0pt}
        \resizebox{\linewidth}{!}{%
        \begin{tabular}{l r r r}
            \toprule
            \textbf{Method} & \textbf{Steps} & \textbf{FID} $\downarrow$ & \textbf{FVD} $\downarrow$ \\
            \midrule
            
            % \multicolumn{4}{l}{\scriptsize \textit{State-of-the-Art Methods}} \\
            % \addlinespace[3pt]
            MagicDrive-V2\cite{gao2025magicdrive} & 30 & 20.9 & 94.8 \\
            DreamForge\cite{mei2024dreamforge} & 30 & 14.6 & 103.6 \\
            Drive-WM\cite{wang2024driving} & 50 & 15.2 & 122.7 \\
            DriveDreamer-2\cite{zhao2025drivedreamer} & - & 11.2 & 55.7 \\
            UniMLVG$^\dagger$\cite{chen2024unimlvg} & 50 & \textbf{5.8} & \underline{36.1} \\
            OpenDWM\cite{opendwm} & 40 & \underline{6.8} & 38.8 \\
            GeoFlow (Ours) & \textbf{15} & \underline{6.8} & \textbf{32.5} \\

            \bottomrule
        \end{tabular}
        }
        \caption{Video Quality Comparison}
        \label{tab:main_results}
    \end{subtable}
    \hfill
    % ================= RIGHT COLUMN =================
    \begin{minipage}[t]{0.44\textwidth}
        \centering
        % -------- TOP SMALL TABLE --------
        \begin{subtable}[t]{\textwidth}
            \centering            
            \setlength{\tabcolsep}{5.0pt}
            \resizebox{\linewidth}{!}{%
            \begin{tabular}{l r}
                \toprule
                \textbf{Base Model} & \textbf{5-step FVD} \\
                \midrule
                OpenDWM-tvae & 204.8\\
                \ \ +GeoFlow(\textcolor{green!65!black}{-62.5\%}) & 76.7\\ 
                OpenDWM-vae & 121.7\\
                \ \ +GeoFlow(\textcolor{green!65!black}{-59.6\%}) & 49.2\\
                UniMLVG & 72.6\\
                \ \ +GeoFlow(\textcolor{green!65!black}{-38.6\%}) & 44.6  \\
                \bottomrule
            \end{tabular}
            }
            \caption{Generality Analysis of GeoFlow}
            \label{tab:generality_analysis}
        \end{subtable}
    \end{minipage}
    % \vspace{-6mm}
\end{table}
\begin{table}[!t]
    \centering
    \small
    \caption{Comparison of FID and FVD metrics across different sampling steps. Geo-Flow consistently outperforms the baseline and achieves better quality with 8 steps than base model with 40 steps in FVD.}
    % \vspace{-2mm}
    \setlength{\tabcolsep}{5.0pt}
    \resizebox{\linewidth}{!}{
    \begin{tabular}{l c c c c c c c c c c c c}
        \toprule
        \multirow{2}{*}{\textbf{Method}} 
        & \multicolumn{2}{c}{\textbf{5-step}} & \multicolumn{2}{c}{\textbf{8-step}} & \multicolumn{2}{c}{\textbf{10-step}} & \multicolumn{2}{c}{\textbf{15-step}} & \multicolumn{2}{c}{\textbf{20-step}} & \multicolumn{2}{c}{\textbf{40-step}} \\
        \cmidrule(lr){2-3} \cmidrule(lr){4-5} \cmidrule(lr){6-7} \cmidrule(lr){8-9} \cmidrule(lr){10-11} \cmidrule(lr){12-13}
        & FID & FVD & FID & FVD & FID & FVD & FID & FVD & FID & FVD & FID & FVD \\
        \midrule
        OpenDWM  & 20.4 & 123.5 & 14.7 & 77.4 & 12.6 & 66.1 & 10.1 & 52.8 & 8.7 & 45.1 & \textbf{6.8} & 38.8 \\
        GeoFlow  & \textbf{11.7} & \textbf{49.2} & \textbf{8.3} & \textbf{38.6} & \textbf{7.5} & \textbf{35.0} & \textbf{6.8} & \textbf{32.5} & \textbf{6.8} & \textbf{32.6} & 6.9 & \textbf{34.0} \\
        \bottomrule
    \end{tabular}
    }
    % \vspace{-5mm}
    \label{tab:step_comparison_full}
\end{table}

\begin{figure}[tbp]
    \centering
    
    % 上子图，占60%高度
    \begin{subfigure}[b]{\linewidth}
        \includegraphics[width=\linewidth, height=0.28\linewidth, keepaspectratio=false]{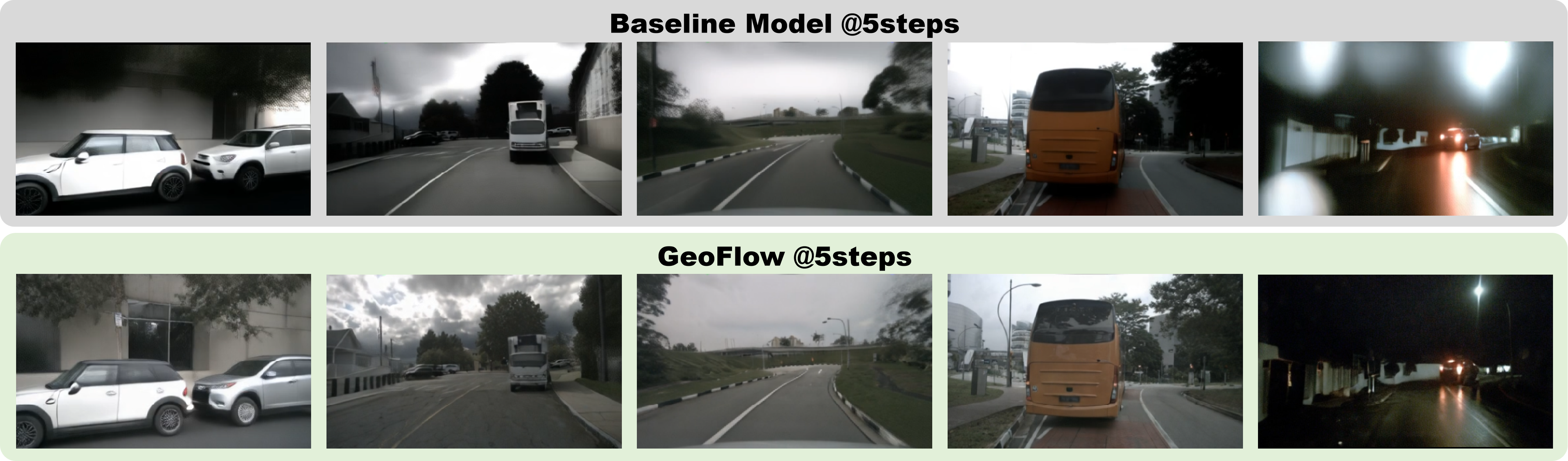}
        \caption{Image quality comparison.}
        \label{fig:image_comparison}
    \end{subfigure}
    
    % \vspace{4pt}  % 调整子图间距
    
    % 下子图，占40%高度
    \begin{subfigure}[b]{\linewidth}
        \includegraphics[width=\linewidth, height=0.32\linewidth, keepaspectratio=false]{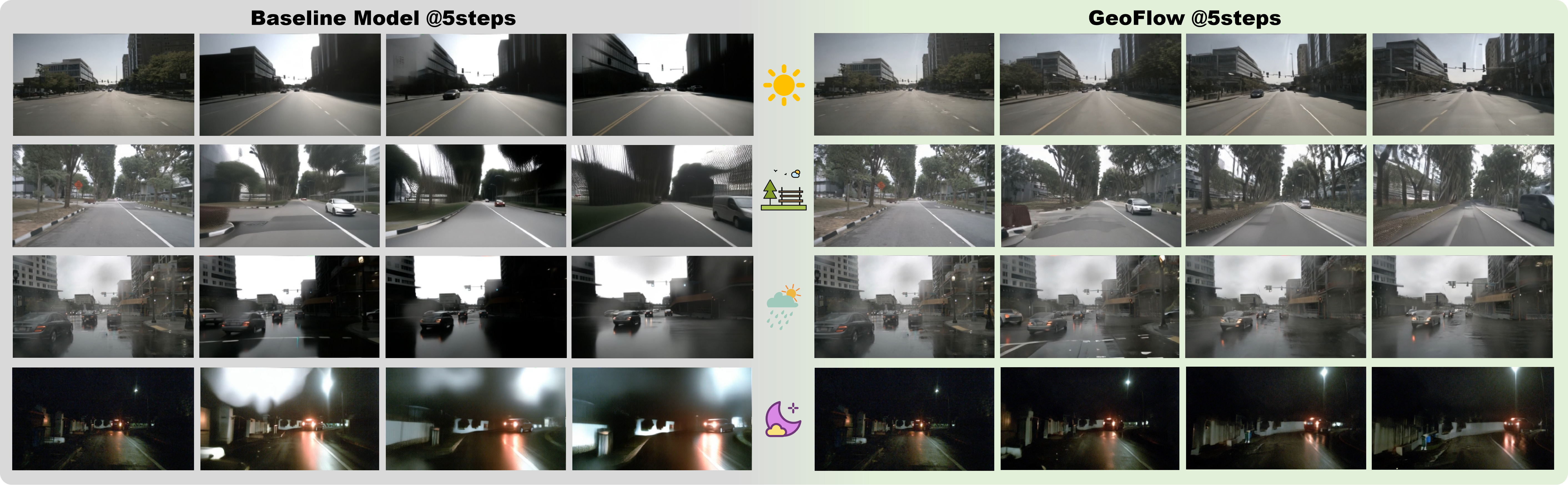}
        \caption{Video temporal consistency comparison.}
        \label{fig:video_comparison}
    \end{subfigure}
    
    \caption{5-step generation visual quality comparison between GeoFlow and Baseline model. All these videos are generated with single reference frame.}
    \label{fig:visual_comparison}
    % \vspace{-6mm}
\end{figure}

\noindent \textbf{Visual Qualitative Comparison.}
Qualitatively, our method produces substantially cleaner and more coherent videos in the few-step regime. In particular, with only \textbf{8 sampling steps}, GeoFlow already yields visual results that are comparable in FVD to the Baseline using \textbf{40 steps}, while the Baseline at few steps often exhibits noticeable blurriness and structural artifacts, as shown in Fig.~\ref{fig:teaser_visual} and Fig.~\ref{fig:visual_comparison}. This demonstrates that our geometry-aligned prior provides a much better initialization, enabling the model to reach a high-quality generation manifold with significantly fewer denoising iterations.

\noindent \textbf{Generality Analysis.}
As demonstrated in Table~\ref{tab:generality_analysis}, our method consistently and significantly improves few-step driving video generation quality across diverse architectures. 
We validate this by experimenting with various models within the OpenDWM codebase, including those equipped with Temporal (3D) VAE and those with Spatial (2D) VAE. The results show that our approach substantially reduces FVD (as shown in \textcolor{green!65!black}{green} percent number) in the few-step inference setting. 
Furthermore, when applied to a stronger base model such as UniMLVG, GeoFlow continues to deliver notable improvements.
% , demonstrating its effectiveness and complementarity to existing advancements. 
This highlights the \textbf{plug-and-play} nature of our approach, offering an efficient path to enhance few-step driving video generators without large computational cost.

% -----------------------------------------------------------------------------
% 4.3 Efficiency Analysis (New Section)
% -----------------------------------------------------------------------------

% \vspace{-0.2cm}
\subsection{Efficiency Analysis}
\label{subsec:efficiency}
% \vspace{-0.2cm}

We provide a comprehensive analysis of the proposed method's efficiency from both training and inference perspectives.

\begin{wrapfigure}[15]{l}{0.4\textwidth} % {位置: r=右, l=左}{宽度}
    \centering
    % \vspace{-0.9cm} % 如果图片这也留白太多，可以打开这行往上提
    \includegraphics[width=\linewidth]{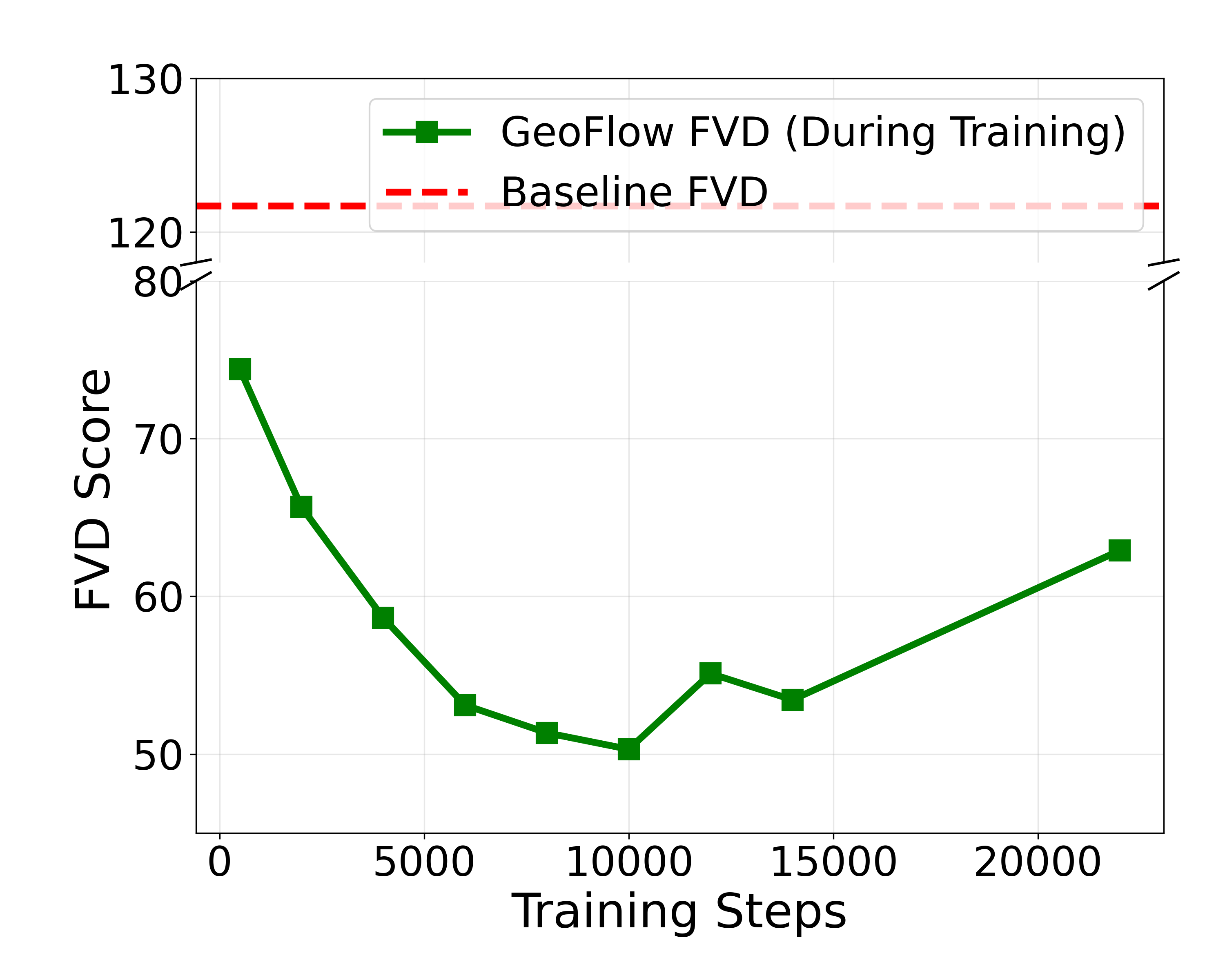}
    % \vspace{-8mm} % 保持你原来的垂直间距调整
    \caption{\textbf{Training Convergence.}}
    % \vspace{-5mm} % 保持你原来的垂直间距调整
    \label{fig:convergence_train}
    % \vspace{-10pt} % 如果下方文字离得太远，可以在这里微调
\end{wrapfigure}

\noindent \textbf{Fast Training Convergence.}
A key advantage of GeoFlow is that it simplifies the learning objective from generating from scratch to residual refinement. 
We argue that the model can efficiently adapt to this new paradigm, because base model has inherently developed such refinement capabilities during its original denoising training phase.
To validate this, we monitor the generation quality (FVD) on the validation set at different training milestones.
As illustrated in Fig.~\ref{fig:convergence_train}, our model exhibits rapid convergence. The FVD score drops significantly within the first thousands iterations (only hundreds of training steps can achieve significant FVD drop in few-step generation) and saturates around 10,000 iterations.
The total adaptive training cost is less than 30 H100 GPU hours. 
This demonstrates that our framework can be easily adapted to existing base models without requiring expensive distillation or fully retraining. 

\noindent \textbf{Inference Efficiency.}
We further evaluate the inference efficiency. As demonstrated in Sec.~\ref{subsec:training}, although our method employs a shorter autoregressive chunk size ($L=6$) compared to the baseline ($L=19$) to ensure geometric validity, the total generation time to achieve SOTA quality is still significantly reduced.
Specifically, generating a 16-frame clip using our method (8 steps $\times$ 4 passes) is \textbf{$4.2\times$} faster than the baseline (40 steps $\times$ 1 pass). This confirms that the overhead of geometric warping is marginal compared to the computational savings from reduced denoising iterations.
Detailed time costs on one L20 GPU are shown in Table~\ref{tab:latency_breakdown}.

\begin{wraptable}[12]{r}{0.45\textwidth} % {r} 表示靠右，{0.45\textwidth} 表示占据 45% 的行宽
    \centering
    % \vspace{-4mm} % 向上微调，使表格顶部与段落第一行对齐
    \caption{Breakdown of 5-step inference latency for a single 6-frame clip.}
    \label{tab:latency_breakdown}
    % \vspace{2mm}
    \small % 缩小表格字体，适应侧边排版
    \begin{tabular}{lrr}
        \toprule
        \textbf{Stage} & \textbf{Time (s)} & \textbf{Prop.} \\
        \midrule
        Geo. Recon.      & 0.43 & 2.84\% \\
        Feat. Rend.     & 0.49 & 3.24\% \\
        ODE Solving     & 13.83 & 91.65\% \\
        \midrule
        \textbf{Total}  & 15.09 & 100\% \\
        \bottomrule
    \end{tabular}
    % \vspace{-8mm} % 减少表格下方多余的空白
\end{wraptable}

\noindent \textbf{Efficiency-Quality Trade-off.}
Table~\ref{tab:step_comparison_full} illustrates the trade-off between video quality and inference steps for the base model and GeoFlow, demonstrating that the GeoFlow framework substantially enhances the generation quality of the base model. 
GeoFlow achieves significant acceleration; specifically, our model with only \textbf{8 sampling steps} attains an FVD of \textbf{38.6}, surpassing the Baseline with \textbf{40 steps} (FVD 38.8). This represents a \textbf{5× steps reduction} without compromising visual quality.
To further visualize the acceleration capability, we plot the FVD scores against the number of sampling steps in Fig.~\ref{fig:trade_off_curve}. 
The Baseline (red curve) exhibits a slow convergence rate, requiring more than 40 steps to reach a saturated performance. In contrast, our method (green curve) starts from a much lower FVD and converges rapidly, achieving high-quality results in around 10 steps. 
% This validates our hypothesis that the geometry-aware prior significantly straightens the probability flow trajectory.

% -----------------------------------------------------------------------------
% 4.4 Ablation Studies
% -----------------------------------------------------------------------------

% \begin{table}[h]
%     \centering
%     \caption{\textbf{Ablation on source distribution strategies.} We compare Naïve Warping, Global Uniform Noise injection, and our Spatially-Adaptive strategy. 'Naïve Warping' suffers from severe artifacts, while 'Global Noise' degrades detail preservation.}
%     \label{tab:ablation_strategy}
%     \begin{tabular}{l|c|cc}
%         \toprule
%         Strategy & Noise Injection & FVD ($\downarrow$) & mIoU ($\uparrow$) \\
%         \midrule
%         Baseline & Standard Gaussian (Pure Noise) & 40.7 & [TODO] \\
%         \midrule
%         Naïve Warping & None & [TODO] & [Low] \\
%         Global Noise & Uniform Mixing & 60.5 & [Mid] \\
%         \textbf{Ours} & \textbf{Spatially-Adaptive} & \textbf{37.9} & \textbf{[High]} \\
%         \bottomrule
%     \end{tabular}
% \end{table}
% \vspace{-0.2cm}
\subsection{Ablation Studies}
\label{subsec:ablation}
% \vspace{-0.1cm}
% We verify the effectiveness of our proposed method by conducting ablation studies on the NuScenes validation set.
% Our ablation studies primarily investigate the impact of different noise injection strategies and various noise mask designs, as well as the selection of metric depth models during inference. 
% Furthermore, we observe that when employing Classifier-Free Guidance (CFG), a smaller guidance scale yields better generation quality. This aligns with our expectation that the velocity predicted by our model does not require strong conditional guidance, as the distribution constructed by GeoFlow already encodes substantial conditional information. Consequently, the model should focus on generating the noise-masked regions with a lower CFG scale, allowing the geometry-aware prior to take effect.
% Unless otherwise specified, all ablation models are trained for the same number of iterations and evaluated using 10 sampling steps.

\begin{figure*}[t]
    \centering
    % ==================================================
    % 左侧列：包含子图 (a)
    % 宽度设置为 0.48\linewidth，留一点空隙给中间
    % ==================================================
    \begin{minipage}[b]{0.43\linewidth}
        \centering
        \begin{subfigure}[b]{\linewidth}
            \centering
            \includegraphics[width=\linewidth]{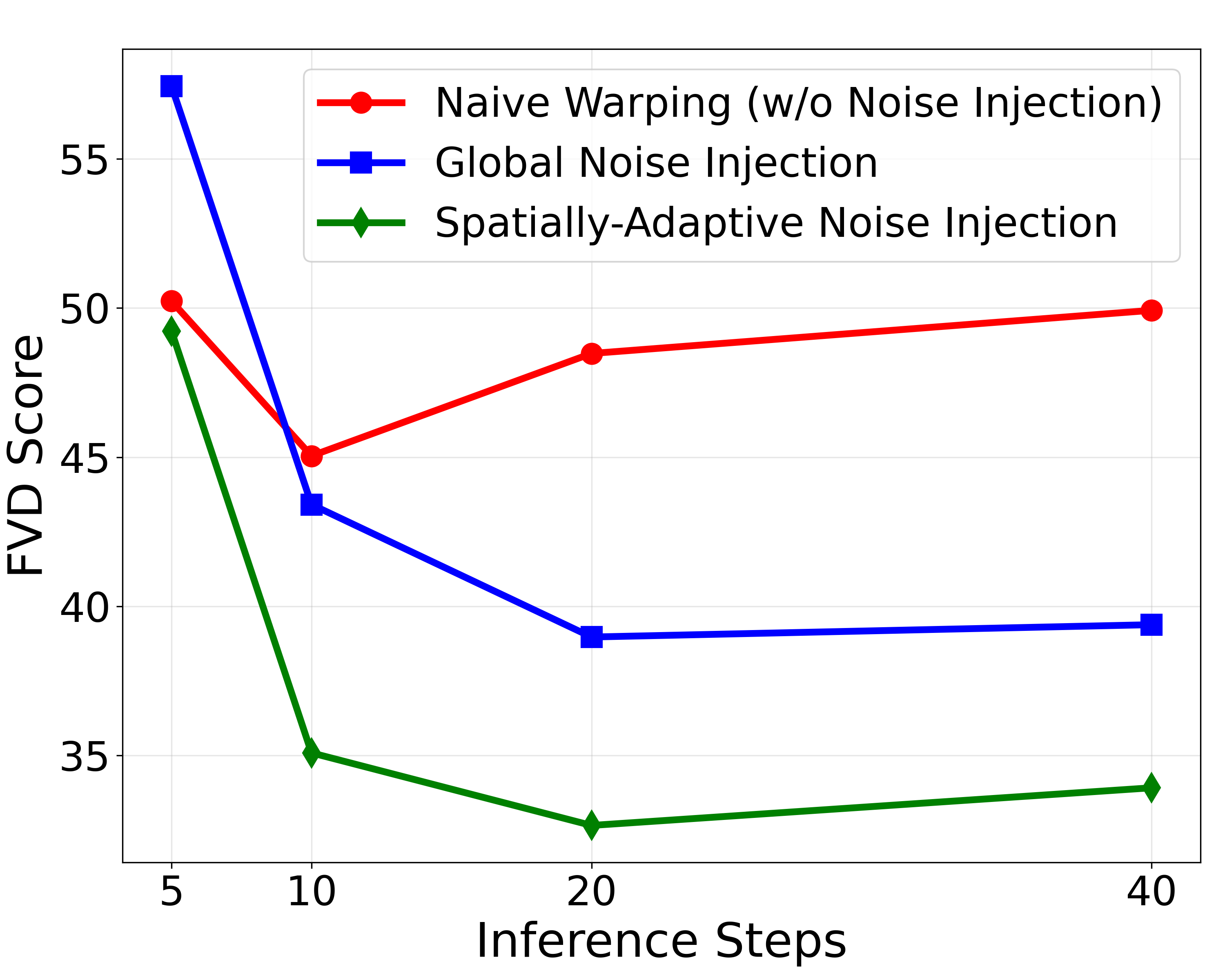}
            % \vspace{-8mm} % [原始 vspace]
            \caption{\textbf{Ablation of Noise Injection.}}
            \label{fig:noise_ablation}
        \end{subfigure}
        % 如果左边图片太短，可以在这里加 \vfill 或者 \vspace 来调整垂直对齐
    \end{minipage}
    \hfill % 左右两列之间的弹性间距
    % ==================================================
    % 右侧列：包含子图 (b) 和 (c)，上下堆叠
    % 宽度设置为 0.48\linewidth
    % ==================================================
    \begin{minipage}[b]{0.55\linewidth}
        \centering    
        % --- 子图 (b) ---
        % \vspace{-6mm}
        \begin{subfigure}[b]{\linewidth}
            \centering
            % \vspace{-0.8cm} % [原始 vspace]
            \includegraphics[width=\linewidth]{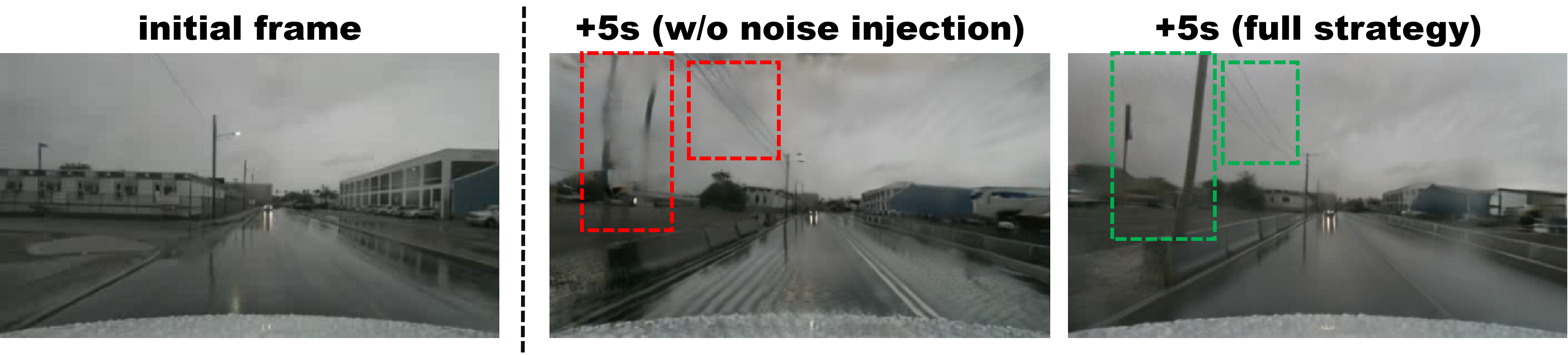}
            % \vspace{-6mm} % [原始 vspace]
            \caption{\textbf{Effectiveness of noise injection.}}
            \label{fig:no_noise_vs_ours}
        \end{subfigure}

        % \vspace{-2mm} % [可选] b和c之间的间距，根据需要打开调整
        % --- 子图 (c) ---
        \begin{subfigure}[b]{\linewidth}
            \centering
            % \vspace{-8mm} % [原始 vspace]
            \includegraphics[width=\linewidth]{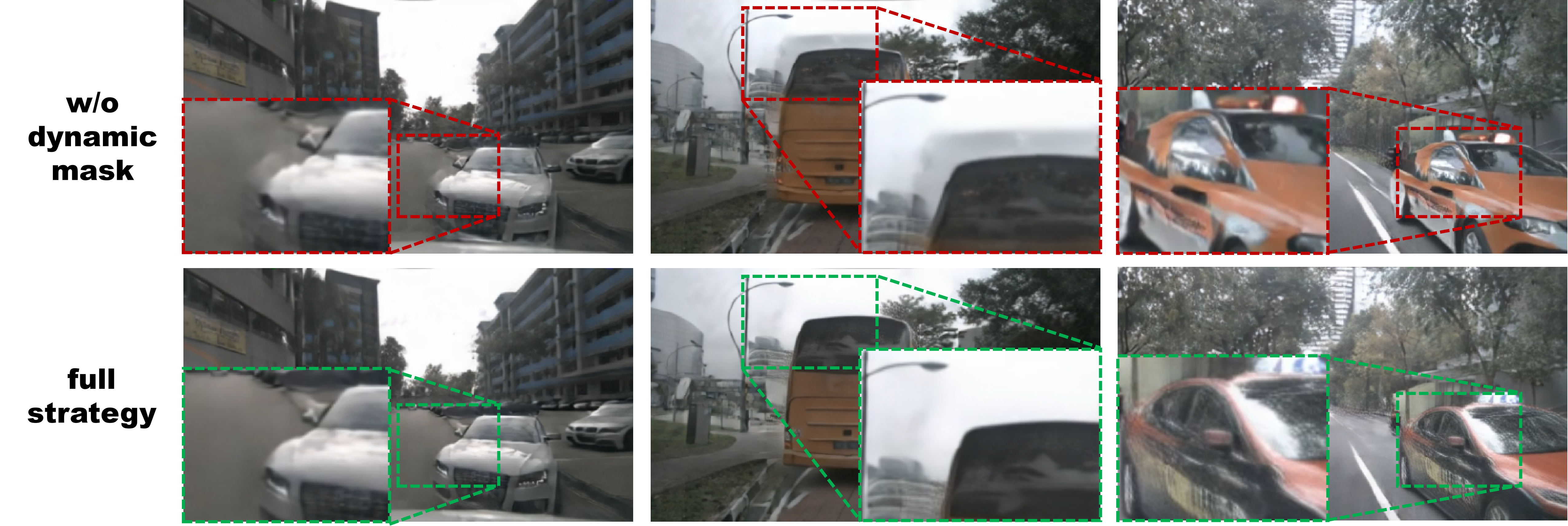}
            % \vspace{-6mm} % [原始 vspace]
            \caption{\textbf{Ablation of dynamic mask.}}
            \label{fig:wo_dyn}
        \end{subfigure}
    \end{minipage}
    % \vspace{-5mm} % [可选] 调整整个大图 caption 与图片的距离
    \caption{\textbf{Qualitative and quantitative ablations.} 
    \textbf{(a)} Quantitative comparison of different injection strategies. 
    \textbf{(b)} Visual comparison of generating with and without noise injection.
    \textbf{(c)} Visual ablation of the dynamic masking strategy.} 
    \label{fig:ablation_main}
    % \vspace{-5mm} % [可选] 调整大图与正文的距离
\end{figure*}

% \begin{wrapfigure}{t}{0.6\textwidth}
%     \centering
%     \vspace{-0.8cm}
%     \includegraphics[width=\linewidth]{figs/no_noise_vs_ours.png}
%     \vspace{-6mm}
%     \caption{\textbf{Ablation of noise injection.} 
%     % Directly using rendered latents without noise injection leads to visible artifacts, while our adaptive noise injection produces cleaner and more consistent results.
%     }
%     \label{fig:no_noise_vs_ours}
%     \vspace{-0.7cm}
% \end{wrapfigure}
% \paragraph{Effectiveness of Noise Injection.}
\noindent \textbf{Effectiveness of Noise Injection.}
\label{para:noise_injection}
Fig.~\ref{fig:noise_ablation} compares different strategies for constructing the source distribution $\hat{x}_0$. 
First, \textbf{Naïve Warping}, which directly uses warped latents without noise injection, results in poor performance. 
As shown in Fig.~\ref{fig:no_noise_vs_ours}, the deterministic nature of flawed geometric priors causes the model to overfit to rendering artifacts (e.g., stretching textures), leading to severe visual distortions. 
% 如图中红色折线所示，由于训练时没有噪声注入，限制了模型对数据流型的探索，导致推理步数多的时候出现误差累积，生成质量反而降低。
Also, as illustrated by the red curve in Fig.~\ref{fig:noise_ablation}, the lack of noise injection during training restricts the model's exploration of the data manifold, leading to error accumulation at higher inference steps and degrading the generation quality.
Second, the addition of \textbf{global uniform noise} mitigates these artifacts to some extent. However, as shown in Fig.~\ref{fig:noise_ablation}, its FVD score remains inferior to our method because uniform noise indiscriminately corrupts perfectly aligned static backgrounds (e.g., road markings), counteracting the benefits of geometric warping by forcing the model to reconstruct valid structures from scratch. 
Finally, our full strategy achieves the best performance by using the proposed reliability mask to selectively inject noise. 
% This strategy preserves the sharp details of static backgrounds while allowing the model to correct geometric errors and hallucinate new content in occluded areas.
% \begin{figure}[t]
%     \centering

%     % \vspace{-0.3cm}
%     \includegraphics[width=\linewidth]{figs/no_noise_vs_ours.png}
%     \caption{\textbf{Ablation study of our noise injection strategy.} Directly using rendered latents without noise injection leads to visible artifacts, while our adaptive noise injection produces cleaner and more consistent results.}
%     \label{fig:no_noise_vs_ours}
%     % \vspace{-0.8cm}
% \end{figure}

% \begin{wrapfigure}{t}{0.6\textwidth}
%     \centering
%     \vspace{-8mm}
%     \includegraphics[width=\linewidth]{figs/wo_dyn.png}
%     \vspace{-6mm}
%     \caption{\textbf{Ablation of dynamic mask.} 
%     % Driving video contains dynamic objects like moving vehicles, while reconstructed point cloud is static; without noise masking of dynamic region, visual artifacts will appear near the vehicle.
%     }
%     \label{fig:wo_dyn}
%     \vspace{-7mm}
% \end{wrapfigure}
\noindent \textbf{Component Analysis of the Noise Injection.}
We further dissect the contribution of each component in our Noise Injection Mask $\mathbf{M}$ in Table~\ref{tab:ablation_components}.
% \textbf{Dynamic Mask ($\mathbf{M}_{dyn}$).} 
Removing $\mathbf{M}_{dyn}$ leads to significant ghosting artifacts. Since the warping operation projects dynamic objects (e.g., moving cars) to incorrect locations based on their past positions, the model struggles to remove these incorrect features without explicit noise injection (as shown in Fig.~\ref{fig:wo_dyn}). 
Introducing the dynamic mask significantly reduces artifacts around moving vehicles and improves the visual quality of vehicles in the generated videos.
% \textbf{Occlusion Mask ($\mathbf{M}_{occ}$).} 
% \begin{figure}[t]
%     \centering
%     % \vspace{-2mm}
%     \includegraphics[width=\linewidth]{figs/wo_dyn.png}
%     \caption{\textbf{Ablation study of dynamic mask.} Driving video contains dynamic objects like moving vehicles, while reconstructed point cloud is static; without noise masking of dynamic region, visual artifacts will appear near the vehicle.}
%     \label{fig:wo_dyn}
%     \vspace{-4mm}
% \end{figure}
Without $\mathbf{M}_{occ}$, the model receives invalid features (e.g., zero initialization for background rendering) in the out-of-view regions. Masking these areas allows the model to perform standard in-painting.
% ensuring seamless transitions between known and unknown regions.
% \textbf{Uncertainty Mask ($\mathbf{M}_{unc}$).} 
% Incorporating depth uncertainty further boosts performance by filtering out geometric noise in challenging areas like the sky or reflective surfaces. 

\noindent \textbf{Robustness to Depth Estimation.}
% 为了验证我们的模型对不同的深度估计模型具有泛化性，我们在推理时选择了和训练时不同的MDE模型。在训练时我们使用MapAnything-1.0，在推理时，我们替换为MapAnything-v1.1，此外不需要任何的代码改动。实验表明（如表x所示），MA1.1有更好的深度估计质量以及更准确的不确定度估计，因此模型在推理步数较少时的性能取得了提升。然而由于不同的MDE模型选择导致的训练推理差异，当步数较多的时候速度方向的微小误差会不断累积，导致视频质量不如MA1.0。
% 我们也验证了GeoFlow并不依赖十分准确的深度估计，即便深度不准确，或者人工在渲染结果中引入扰动，GeoFlow仍然能显著提升少步数下的视频生成质量。具体结果见附录。
We further verify the robustness of our model by substituting the training used Metric Depth Estimation model (Map-Anything-1.0) with MapAnything-v1.1 and DepthAnything-3 during inference in a zero-shot manner. 
As demonstrated in Table~\ref{tab:MDE_selection}, the enhanced depth and uncertainty estimation of MapAnything-v1.1 improves video generation performance at low sampling steps. 
% Nevertheless, the train-inference discrepancy between the MDE models causes minor velocity errors that accumulate at higher steps, leading to suboptimal quality compared to the original setup.
Replacing MapAnything with DepthAnything-3 model can also achieve SOTA performance.
Despite this, we show that GeoFlow can boost few-step generation quality even with inaccurate depth or artificially perturbed rendering results, as detailed in the supplementary material.
% \vspace{-0.3cm}

\begin{table}[h]
    \centering
    % ========== 左侧表格 ==========
    \begin{minipage}[t]{0.48\textwidth}
        \centering
        % \vspace{-15mm}
        \setlength{\tabcolsep}{4pt}
        \caption{\textbf{Component analysis of Noise Injection Masks.} Inference step is set to 10.}
        % \vspace{-2mm}
        \resizebox{\linewidth}{!}{%
        \begin{tabular}{ccc|cc}
            \toprule
            $\mathbf{M}_{occ}$ & $\mathbf{M}_{unc}$ & $\mathbf{M}_{dyn}$ & FID ($\downarrow$) & FVD ($\downarrow$) \\
            \midrule
            - & - & - & 8.2 & 43.6 \\
            \checkmark & - & - & 7.9 & \underline{40.0} \\
            \checkmark & \checkmark & - & \underline{7.6} & 40.2 \\
            \checkmark & \checkmark & \checkmark & \textbf{7.5} & \textbf{35.0} \\
            \bottomrule
        \end{tabular}
        }
        \label{tab:ablation_components}
    \end{minipage}%
    \hfill
    % ========== 右侧图片 ==========
    % \begin{minipage}[t]{0.49\textwidth}
    %     \centering
    %     % \vspace{-13.2mm}
    %     \setlength{\tabcolsep}{2.5pt}
    %     \caption{\textbf{Robustness analysis of MDE model selection.} MapAnything-v1.1 is used in a zero-shot manner.}
    %     \vspace{-2mm}
    %     \begin{tabular}{c c c c c}
    %         \toprule
    %         \multirow{2}{*}{\textbf{Steps}} 
    %         & \multicolumn{2}{c}{\textbf{MA-v1.0}} & \multicolumn{2}{c}{\textbf{MA-v1.1}} \\
    %         \cmidrule(lr){2-3} \cmidrule(lr){4-5}
    %         & FID($\downarrow$) & FVD($\downarrow$) & FID($\downarrow$) & FVD($\downarrow$) \\
    %         \midrule
    %         5   & 11.7 & 49.2 & 11.1 & 44.1\\
    %         10  & 7.5 & 35.0 & 7.9 & 33.1 \\
    %         % 15  & 6.8 & 32.5 & 7.8 & 62.4 \\
    %         40  & 6.9 & 34.0 & 7.8 & 35.6 \\
    %         \bottomrule
    %     \end{tabular}
    %     \label{tab:MDE_selection}
    % \end{minipage}
    \begin{minipage}[t]{0.48\textwidth}
        \centering
        % \vspace{-13.2mm}
        \setlength{\tabcolsep}{4.2pt}
        \caption{\textbf{Robustness analysis of depth model selection.} The two models are introduced in a zero-shot manner.}
        % \vspace{-2mm}
        \resizebox{\linewidth}{!}{%
        \begin{tabular}{c c c c c}
            \toprule
            \multirow{2}{*}{\textbf{Steps}} 
            & \multicolumn{2}{c}{\textbf{MA-v1.1}} & \multicolumn{2}{c}{\textbf{DA-3}} \\
            \cmidrule(lr){2-3} \cmidrule(lr){4-5}
            & FID($\downarrow$) & FVD($\downarrow$) & FID($\downarrow$) & FVD($\downarrow$) \\
            \midrule
            5   & 11.1 & 44.1 & 13.1 & 55.1\\
            10  & 7.9 & 33.1 & 8.0 & 38.2 \\
            40  & 7.8 & 35.6 & 7.2 & 37.3 \\
            \bottomrule
        \end{tabular}
        }
        \label{tab:MDE_selection}
    \end{minipage}
\end{table}
\section{Conclusion}
% \vspace{-0.1cm}
In this work, we present GeoFlow, a plug-and-play framework that achieves efficient driving video generation by bridging the distributional gap between the source and target distributions. 
By replacing the non-informative standard Gaussian with a Geometry-Aligned Prior Distribution (GAP), we successfully shorten the sampling trajectory, enabling Flow Matching models to generate high-fidelity driving videos in just several sampling steps.
Crucially, GeoFlow achieves this efficiency with low adapting cost and no architectural changes, offering a practical and scalable solution for high-throughput data synthesis in autonomous driving.

\section*{Acknowledgements}
In this work, we are supported by the National Natural Science Foundation of China 62372029, 62276016.

% \clearpage\mbox{}Page \thepage\ of the manuscript.
% \clearpage\mbox{}Page \thepage\ of the manuscript.
% \clearpage\mbox{}Page \thepage\ of the manuscript.
% \clearpage\mbox{}Page \thepage\ of the manuscript.
% \clearpage\mbox{}Page \thepage\ of the manuscript. This is the last page.
% Now we have reached the maximum length of an ECCV \ECCVyear{} submission (excluding references).
% References should start immediately after the main text, but can continue past p.\ 14 if needed.
% \clearpage  % TODO REVIEW/FINAL: This \clearpage needs to be removed from both review and camera-ready versions.

\bibliographystyle{splncs04}
\bibliography{main}

\clearpage
% \section*{\centering Supplementary Material}
% \subfile{chapters/appendix}

\end{document}